\documentclass[sn-apa]{sn-jnl}%

\usepackage{graphicx}%
\usepackage{multirow}%
\usepackage{amsmath,amssymb,amsfonts}%
\usepackage{amsthm}%
\usepackage{mathrsfs}%
\usepackage[title]{appendix}%
\usepackage{xcolor}%
\usepackage{textcomp}%
\usepackage{manyfoot}%
\usepackage{booktabs}%
\usepackage{algorithm}%
\usepackage{algorithmicx}%
\usepackage{algpseudocode}%
\usepackage{listings}%
\usepackage{booktabs, 
            multirow, threeparttable}   %
\usepackage{siunitx}                    %
\usepackage{xparse}                     %
\usepackage{makecell}
\usepackage{subcaption}
\usepackage{hyperref}
\usepackage{float}

\theoremstyle{thmstyleone}%

\theoremstyle{thmstyletwo}%

\theoremstyle{thmstylethree}%

\begin{document}

\title[]{Your Next State-of-the-Art Could Come from Another Domain: A Cross-Domain Analysis of Hierarchical Text Classification}

\author*[1]{\fnm{Nan} \sur{Li}}\email{nan.li@ugent.be}

\author[1]{\fnm{Bo} \sur{Kang}}\email{bo.kang@ugent.be}

\author[1]{\fnm{Tijl} \sur{De Bie}}\email{tijl.debie@ugent.be}

\affil[1]{\orgdiv{IDLab}, \orgname{Ghent University}, \orgaddress{\city{Ghent}, \postcode{9052}, \country{Belgium}}}

\abstract{Text classification with hierarchical labels is a prevalent and challenging task in natural language processing. Examples include assigning ICD codes to patient records, tagging patents into IPC classes, assigning EUROVOC descriptors to European legal texts, and more. Despite its widespread applications, a comprehensive understanding of state-of-the-art methods across different domains has been lacking.
In this paper, we provide the first comprehensive cross-domain overview with empirical analysis of state-of-the-art methods. We propose a unified framework that positions each method within a common structure to facilitate research. Our empirical analysis yields key insights and guidelines, confirming the necessity of learning across different research areas to design effective methods. Notably, 
under our unified evaluation pipeline, we achieved new state-of-the-art results by applying techniques beyond their original domains.}


\maketitle

\section{Introduction}\label{sec:intro}

Text classification with hierarchical labels is a fundamental challenge in natural language processing, where the goal is to assign one or more labels from a hierarchically-organized label set to each text input. This problem appears across diverse domains, such as medical coding that assigns International Classification of Diseases (ICD) codes to patient records \citep{edin2023automated}, patent classification that predicts International Patent Classification (IPC) codes for patent documents \citep{kamateri2024will}, and extreme multi-label classification tasks in legal, wikipedia, and e-commerce domains \citep{Bhatia16extreme}. While these applications may seem distinct, they share a common core: classifying text with labels that have inherent hierarchical relationships.

Despite this commonality, current research is largely confined within individual domains. Methods are typically developed and evaluated only on domain-specific datasets, with minimal cross-domain analysis or comparison. This domain-centric approach has led to three significant knowledge gaps in the research literature: (1) how methods that excel in one domain compare to domain-specific approaches in others; (2) which architectural components (to which we refer as \emph{submodules}, such as text encoders, label encoders, and prediction mechanisms) are truly domain-specific versus universally effective; and (3) how dataset characteristics, rather than domain origin, influence method performance.

To address these gaps, we present the first comprehensive cross-domain analysis of hierarchical text classification. Our work makes several key contributions:
\begin{itemize}
    \item \textbf{Cross-Domain Survey:} We present the first comprehensive cross-domain survey of hierarchical text classification methods, dissecting 32 representative methods and abstracting common elements that bridge different domains. 
    
    \item \textbf{Unified Framework:} We propose the first domain agnostic framework that decomposes hierarchical text classification methods into nine essential submodules, enabling systematic comparison of approaches across different domains. This framework particularly highlights how different methods utilize hierarchical label information.

    \item \textbf{Cross-Domain Analysis:} We conduct the first large-scale cross-domain evaluation of hierarchical text classification methods (with necessary re-implementations and unified evaluation codebase), analyzing eight state-of-the-art methods across five domains on eight curated datasets. 
    
    \item \textbf{Enhanced Datasets:} We make several dataset contributions: (1) a cleaned version of EurLex with recovered taxonomy and restored original text; (2) a new dataset derived from EurLex, EurLex-DC-410, labeled with directory codes; and (3) a carefully curated selection of eight datasets across five domains, with necessary adaptations to enable fair cross-domain evaluation, publicly available at \url{https://github.com/aida-ugent/cross-domain-HTC}. 
    
    \item \textbf{New State-of-the-Art Results:} We achieved new state-of-the-art results\footnote{These state-of-the-art results are based on our unified evaluation framework. While we carefully reproduced previous methods following their original implementations, some results may differ from those reported in original papers due to differences in preprocessing, evaluation settings, or implementation details.} on NYT-166, SciHTC-83, USPTO2M-632, and MIMIC3-3681 by applying methods from other domains or combining submodules across domains.

    \item \textbf{Empirical Insights:} Our findings through extensive experiments reveal that dataset characteristics and architectural design choices, rather than domain specificity, primarily determine method effectiveness. We encourage researchers and practitioners to look beyond their specific domains, as state-of-the-art performance can often be achieved by leveraging methods and innovations from other fields.

\end{itemize}

The rest of the paper is organized as follows: Section~\ref{sec:related} reviews existing surveys and benchmarks, highlighting the need for cross-domain analysis. Section~\ref{sec:overview} presents our unified framework and formal problem definition, analyzing how different methods utilize label hierarchical information. Section~\ref{sec:evaluation} details our cross-domain evaluation, including dataset selection, method descriptions, and experimental results. Section~\ref{sec:discussion} provides in-depth analysis of our findings, discussing the cross-domain performances of different methods, impact of several key factors and lessons for practitioners. Section~\ref{sec:conclusion} summarizes key findings and discusses the limitations and future research directions.

\section{Related Work}\label{sec:related}
Section~\ref{sec:surveys} presents the surveys and benchmarks for the general problem of multi-label text classification, while Section~\ref{sec:domain-surveys} focuses on domain-specific surveys and benchmarks.

\subsection{General surveys/benchmarks}\label{sec:surveys}
An overview of multi-label text classification is provided by \cite{chen2022survey}, which systematically categorizes methods by data efficiency, feature extraction, and label correlation modeling, but focuses on flat classification without exploring hierarchical structures or cross-domain analysis. An empirical comparison across benchmark datasets is presented in \cite{bogatinovski2022comprehensive}, examining how model performance relates to dataset characteristics, but primarily examines non-textual domains with small label spaces. The work of \cite{wei2022survey} focuses on extreme multi-label learning challenges and collects datasets and tools, but does not address hierarchical structures or provide empirical validation. Similarly, \cite{Bhatia16extreme} maintains a widely cited repository of extreme multi-label classification methods and datasets spanning e-commerce, wikipedia and legal domains, though many datasets only contain preprocessed feature vectors without original texts, limiting their applicability for pretrained language models.

Various aspects of multi-label classification are covered in recent surveys: multiple modalities \citep{han2023survey}, label imbalance \citep{tarekegn2021review}, and deep learning architectures \citep{tarekegn2024deep}, but none comprehensively address text datasets, hierarchical structures, or cross-domain analysis. A review of hierarchical multi-label text classification methods is presented by \cite{liu2023recent}, categorizing them into tree-based, embedding-based and graph-based approaches, but it covers limited domains and lacks empirical validation.

Most recently, \cite{bertalis2024hierarchical} compares hierarchical text classification (HTC) and extreme multi-label classification (XML), evaluating two methods on three datasets from each of these domains. While their observation about XML methods' adaptability to HTC tasks aligns with one aspect of our findings, our work provides the first comprehensive cross-domain analysis spanning five domains, evaluating eight state-of-the-art methods on eight diverse datasets, with a unified framework for analyzing method components and in-depth empirical analyses of their effectiveness.

\subsection{Domain-specific surveys/benchmarks}\label{sec:domain-surveys}
For automatic medical coding, \cite{edin2023automated} benchmarks state-of-the-art models on cleaned MIMIC datasets, providing an open-source evaluation pipeline. For patent classification, \cite{kamateri2024will} surveys classification complexities, reviews key datasets, and evaluates Deep Learning and large language models' potentials, highlighting domain-specific challenges. For legal documents, \cite{chalkidis2019extreme} establishes a benchmark by introducing EurLex-57K and conducting comparisons of several neural architectures, demonstrating that BiGRU-based models consistently outperform CNN-based approaches on legal text classification tasks.

To summarize, while these surveys provide excellent domain-specific insights, our work differs by offering the first cross-domain survey and benchmark evaluation, encompassing a comprehensive coverage of datasets and methods.

\section{A cross-domain overview of text classification with hierarchical labels}\label{sec:overview}
In this section, we provide the first cross-domain overview of text classification with hierarchical labels. We begin with a formal problem formulation in Section~\ref{sec:problem}. Then, Section~\ref{sec:all-methods} presents recent methods from different domains. We analyze these methods within our proposed unified framework (Section~\ref{sec:submodules}), focusing on their common submodules (architectural components) and their utilization of label information.

\subsection{Problem formulation}\label{sec:problem}
Conceptually, text classification with hierarchical labels is the task of assigning one or more relevant labels to a given text input, where these labels are organized in a hierarchical structure. The challenge lies both in understanding the text content and utilizing the relations between labels. The labeling scheme can vary across datasets: some use leaf-only labels (only the most specific labels), some require complete paths (including all ancestors), and some allow partial paths (labels at any level). The prediction task should match the labeling scheme of the dataset.\footnote{For example, given a hierarchy where A.1.1 is a leaf node under A.1, which is under A, a document could be labeled as: (1) {A.1.1} in a leaf-only scheme, (2) {A, A.1, A.1.1} in a full-path scheme, or (3) {A, A.1.1} in a flexible scheme. For simplicity, we do not consider the impact of these labeling schemes on data preprocessing or model design.}

Formally, the goal of text classification with hierarchical labels is to learn a function $f: X \rightarrow Y^*$, where $X$ is the sample space of all possible text sequences and $Y^*$ is the power set of $L$ (the label set), so that for any given sample $x \in X, f(x)$ outputs one or more labels most relevant to $x$. 
The label space can be organized in a hierarchy, defined as a tuple $(L, E, \rho)$ with $L$ being a set of labels, $E \subseteq L \times L$ being a set of edges forming a directed acyclic graph (though mostly a tree), and $\rho \in L$ being a unique root node. 
The hierarchy $(L, E, \rho)$ satisfies: (1) there are no cycles, (2) each node except $\rho$ has at least one parent, and (3) $\rho$ has no parents.

This hierarchical structure distinguishes the task from flat multi-label classification, where labels have no structures to be exploited. The hierarchy can be either given as a taxonomy or learned from the training data via label co-occurrence patterns or other features. The hierarchy provides additional information about label relationships that can be exploited during model training and inference, though the extent and manner of utilizing this information varies across methods.

\subsection{Method selection}\label{sec:all-methods}
We analyzed 32 representative methods published between 2019 and 2024 across multiple domains. While an exhaustive review is beyond our scope, the selected methods encompass diverse approaches and variations, enabling us to identify common patterns and key differences, and to ultimately develop a unified framework.

We started with the recent surveys in Section~\ref{sec:related} and collected the frequently cited papers, following their citations to find their most related predecessors. We further searched papers published after 2023 from Google Scholar using keywords ``hierarchical text classification'', ``multi-label text classification'', ``patent classification'', ``medical coding'' and ``extreme multi-label classification''. We also included papers that cited the benchmark datasets from Google Scholar and PapersWithCode. To limit the scope, we exclude methods published before 2019, unsupervised methods, and those primarily focused on non-textual datasets.

\begin{figure}[!h]
    \centering
    \begin{subfigure}{\textwidth}
        \centering
        \includegraphics[width=\textwidth]{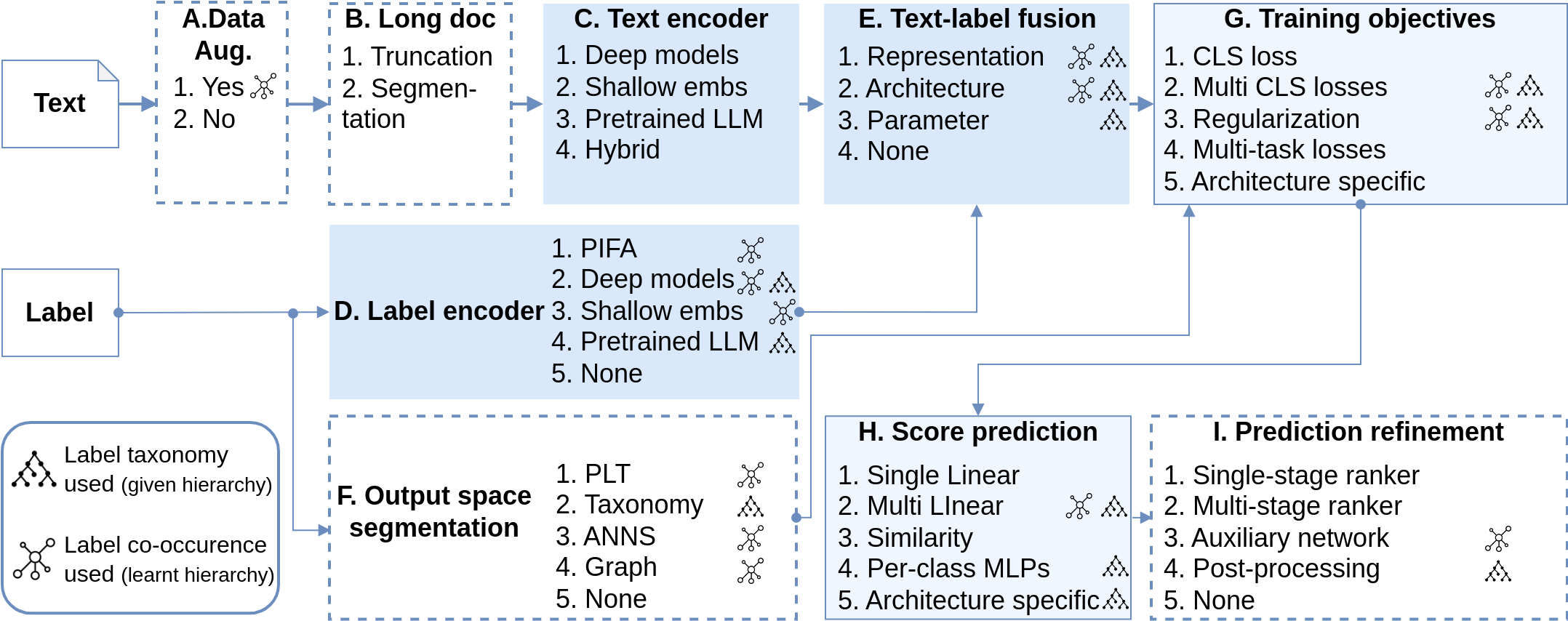}
        \caption{The unified framework with submodules for text classification with hierarchical labels. \\The pictogram \includegraphics[width=1em]{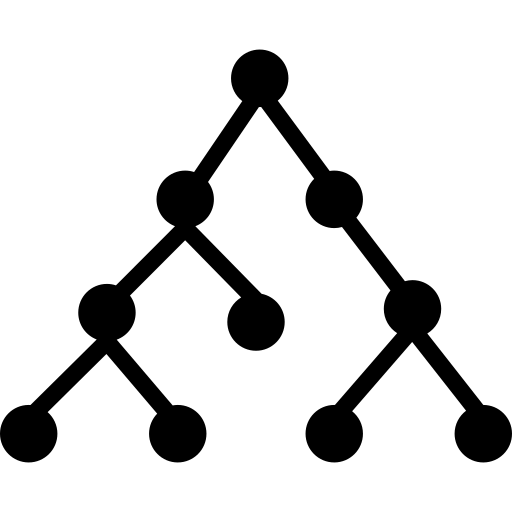} represents that methods use the given taxonomy directly in the corresponding submodule, and \includegraphics[width=1em]{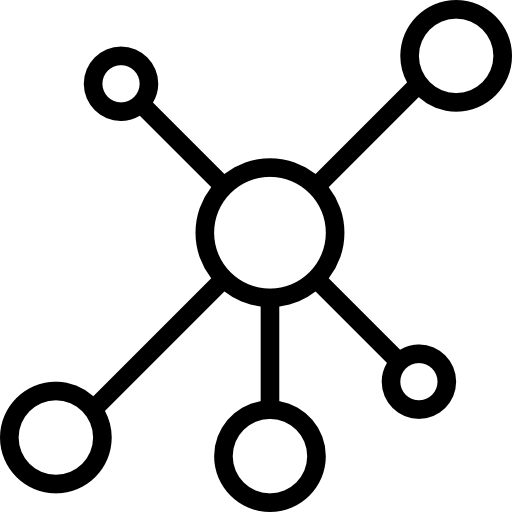} represents that methods learn a label structure from the data via co-occurrence or other features and use that information in the corresponding submodule. \\Abbreviations: \emph{Data Aug.}: data augmentation, \emph{Long Doc}: long document handling, \emph{LLM}: large language model, \emph{CLS loss}: classification loss, e.g. binary cross-entropy, \emph{PIFA}: positive instance feature aggregation, \emph{PLT}: probabilistic label trees, \emph{ANNS}: approximate nearest neighbor search.}   
        \label{fig:submodules}
    \end{subfigure}
    \begin{subfigure}{\textwidth}
        \centering
        \includegraphics[width=\textwidth]{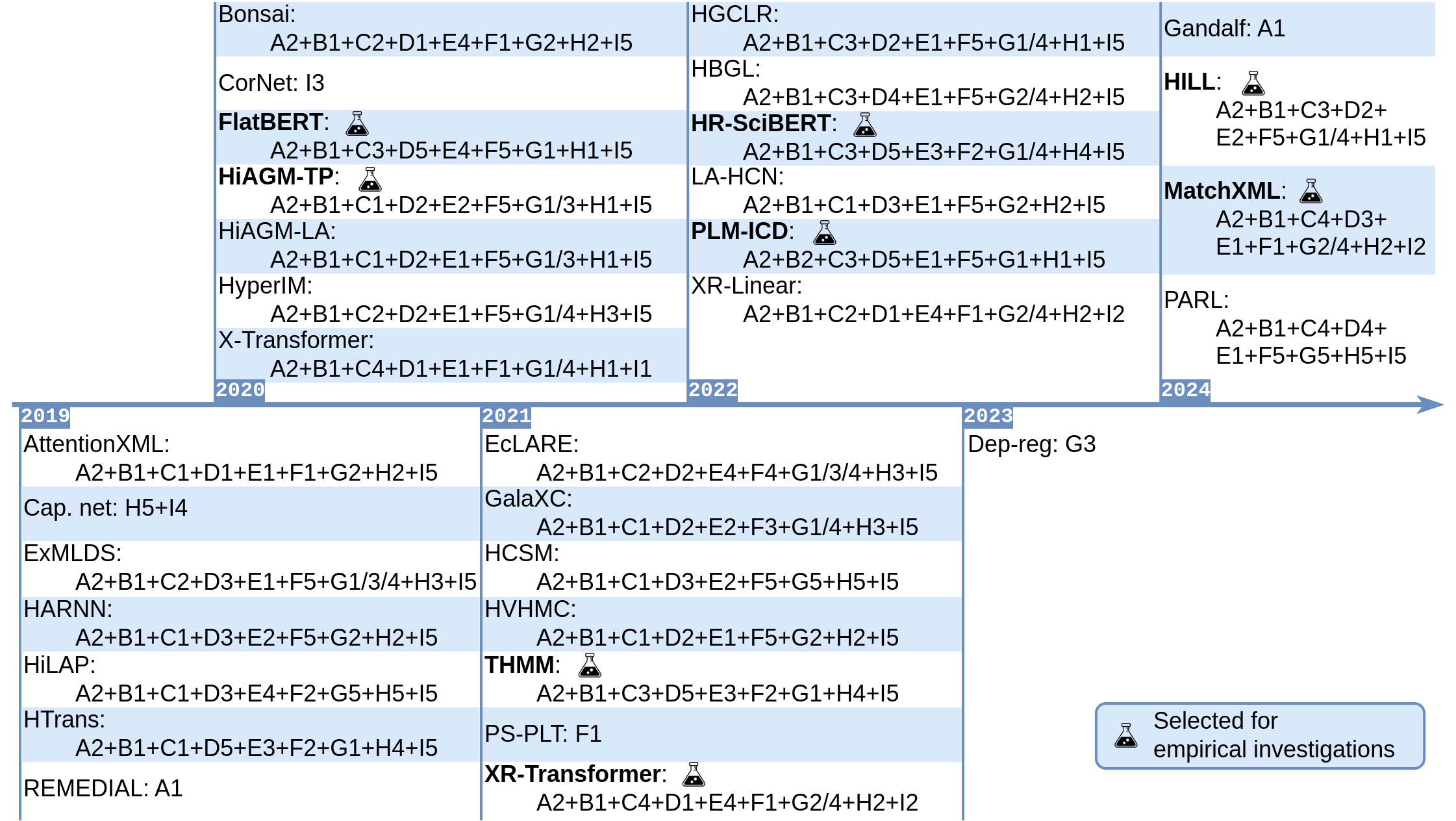}
        \caption{Methods with submodule combinations in chronological order. Each color block represents a method, followed by the combinations of its submodule design choices, where capital letters with numbers (e.g., C1, D2) indicate specific design choices within each submodule defined in Figure~\ref{fig:submodules}, for example, C2 represents shallow embeddings in the text encoder module.}
        \label{fig:methods-timeline}
    \end{subfigure}
    \caption{Our framework and methods with submodule combinations.}
\end{figure}

\subsection{A unified framework of common submodules}\label{sec:submodules}
From the surveyed methods, we abstract a framework constituted by commonly seen architectural components, shown as the nine submodules in Figure~\ref{fig:submodules}. The framework \emph{begins} with optional preprocessing steps: data augmentation (\hyperref[par:data-augmentation]{A}) and long document handling (\hyperref[par:long-document-handling]{B}). 
The \emph{core processing pipeline} starts with parallel encoding of text and labels: a text encoder (\hyperref[par:text-encoder]{C}) transforms input text into embeddings using shallow, deep, or pre-trained models, while a label encoder (\hyperref[par:label-encoder]{D}) converts discrete labels into continuous representations, optionally incorporating hierarchical information. These representations are then integrated through text-label information fusion (\hyperref[par:text-label-fusion]{E}), which can occur at representational, architectural, or parameterization levels. 
For large label spaces, \emph{output space segmentation} (\hyperref[par:output-space]{F}) may be employed for efficiency. The training process is guided by various training objectives (\hyperref[par:training-obj]{G}), including classification losses and other custom objectives. 
Finally, the model makes \emph{inferences} through score prediction (\hyperref[par:score-pred]{H}), followed by optional prediction refinement (\hyperref[par:pred-refine]{I}) to enhance raw outputs using label correlations or ranking mechanisms.

Utilizing our proposed framework, we locate each method\footnote{HiAGM has two variants that differ in text-label fusion, and we select HiAGM-TP for benchmark evaluation. We rename PatentBERT to FlatBERT to reflect its architecture that processes labels without hierarchical information.} by their specific instantiations of the submodules in the framework as shown in Figure~\ref{fig:methods-timeline}.
We further emphasize that label hierarchy in the current problem setting can refer to either a given taxonomy or a learned structure, meaning that the usage of label hierarchy can be explicit (using the given taxonomy directly) or implicit (learning a label structure from the data via co-occurrence or other features). Since not all submodules in existing methods utilize label hierarchy information, we explicitly note such usage both in the following discussion and in Figure~\ref{fig:submodules}.

We now discuss each submodule in detail, highlighting how different methods implement them and utilize label hierarchical information when applicable.

\paragraph{A. Data Augmentation}\label{par:data-augmentation}
Two methods address data imbalance through label information: REMEDIAL \citep{charte2019dealing} balances label distribution by decoupling frequent and rare label co-occurrences using the SCUMBLE metric \citep{charte2014concurrence}, while Gandalf \citep{kharbanda2024gandalf} leverages label co-occurrence graphs to generate soft targets as additional training data.

\noindent\emph{Label info utilization}: Current methods only utilize label co-occurrence patterns.

\paragraph{B. Long document handling}\label{par:long-document-handling}
While most methods simply truncate long documents, PLM-ICD \citep{huang2022plm} employs a more sophisticated approach by segmenting text into smaller parts with pre-defined lengths for later information aggregation.

\noindent\emph{Label info utilization}: Current methods do not explicitly use label information here.

\vspace{1em}
\noindent\begin{minipage}[!t]{\textwidth}
    \paragraph{C. Text encoder} 
    This essential submodule transforms text into embeddings for  input. It has evolved from shallow embeddings to deep learning models, and finally to pre-trained LLMs.

    \vspace{0.5em}
    \label{par:text-encoder}
    \setlength{\tabcolsep}{3pt}
    {\footnotesize
    \begin{tabular}{@{}p{0.25\textwidth}@{}p{0.75\textwidth}@{}}
    \toprule
    \multirow{2}{*}{Shallow embeddings} & TF-IDF features: Bonsai \citep{khandagale2020bonsai}, ECLARE \citep{mittal2021eclare}, ExMLDS \citep{gupta2019distributional} \\
    & Hyperbolic embeddings: HyperIM \citep{chen2020hyperbolic} \\
    \midrule
    \multirow{6}{*}{Deep learning models} & CNN: HiLAP \citep{mao2019hierarchical} \\
    & LSTM: AttentionXML \citep{you2019attentionxml}, HARNN \citep{huang2019hierarchical}, LA-HCN \citep{zhang2022hcn} \\
    & GRU: HTrans \citep{banerjee2019hierarchical} \\
    & Hybrid CNN-RNN: HiAGM \citep{zhou2020hierarchy}, HCSM \citep{wang2021cognitive} \\
    & GCN: HVHMC \citep{xu2021hierarchical} \\
    & Fixed pre-trained embeddings: GalaXC \citep{saini2021galaxc} \\
    \midrule
    Pre-trained LLMs & BERT-family: HILL \citep{zhu2024hill}, PLM-ICD \citep{huang2022plm}, PatentBERT \citep{lee2020patent}, THMM \citep{pujari2021multi}, HGCLR \citep{wang2022incorporating}, HBGL \citep{jiang2022exploiting} \\
    \midrule
    \multirow{2}{*}{Hybrid approaches} & LLM + TF-IDF: X-Transformer \citep{chang2020taming}, XR-Transformer \citep{zhang2021fast}, MatchXML \citep{ye2024matchxml} \\
    & LLM + BM25/GraphSAGE: PARL \citep{li2024scalable} \\
    \botrule
    \end{tabular}}
    \end{minipage}\par
    \vspace{0.5em}

\noindent\emph{Label info utilization}: Label information is not directly used in this module, as we separate text-label fusion into its own submodule.

\vspace{1em}
 
 \noindent\begin{minipage}[t]{\textwidth}
    \paragraph{D. Label encoder} This submodule transforms discrete labels into embeddings. Methods vary from implicit relationship modeling (PIFA) to explicit structure encoding (GNNs), with LLM-based approaches offering a middle ground through semantic understanding.

    \vspace{0.5em}
    \label{par:label-encoder}
    \setlength{\tabcolsep}{3pt}
    \footnotesize
    \begin{tabular}{@{}p{0.25\textwidth}@{}p{0.75\textwidth}@{}}
    \toprule
    \parbox[c]{0.25\textwidth}{\vspace*{\fill}Positive Instance \\Feature Aggregation\vspace*{\fill}} & XR-Transformer \citep{zhang2021fast}, X-Transformer \citep{chang2020taming}, AttentionXML \citep{you2019attentionxml}, Bonsai \citep{khandagale2020bonsai} \\
    \midrule
\end{tabular}
\end{minipage}

\noindent(\textit{Continued on next page.})  %

\noindent\begin{minipage}[t]{\textwidth}
\setlength{\tabcolsep}{3pt}
\footnotesize
\begin{tabular}{@{}p{0.25\textwidth}@{}p{0.75\textwidth}@{}}
\midrule
    \multirow{2}{*}{Shallow embeddings} & Label2Vec: MatchXML \citep{ye2024matchxml}, HCSM \citep{wang2021cognitive}, ExMLDS \citep{gupta2019distributional} \\
    & Hyperbolic embeddings: HyperIM \citep{chen2020hyperbolic} \\
    \midrule
    \multirow{2}{*}{Deep learning models} & Graph neural networks: HiAGM \citep{zhou2020hierarchy}, HVHMC \citep{xu2021hierarchical}, HGCLR \citep{wang2022incorporating}, GalaXC \citep{saini2021galaxc}, ECLARE \citep{mittal2021eclare}, HGCLR \citep{wang2022incorporating} \\
    & Tree-based models: HiAGM \citep{zhou2020hierarchy}, HILL \citep{zhu2024hill} \\
    \midrule
    \multirow{2}{*}{Pre-trained LLMs} & Label text encoding: PARL \citep{li2024scalable} \\
    & Custom pretraining: HBGL \citep{jiang2022exploiting} \\
    \botrule
    \end{tabular}

    \footnotesize
    \end{minipage}\par
    \vspace{0.5em}
    \noindent\emph{Label info utilization}: Label hierarchies are either directly encoded through specialized architectures (GNNs, tree models) or used to guide the pre-training process.
    
\vspace{1em}
        
\noindent\begin{minipage}[t]{\textwidth}
    \paragraph{E. Text-Label information fusion} This submodule integrates text and label information at three distinct levels. While simple concatenation is the most straightforward approach, learnable fusion mechanisms are more popular since they can adapt to different types of label hierarchies.
    
    \vspace{0.5em}
    \label{par:text-label-fusion}
    \setlength{\tabcolsep}{3pt}
    \footnotesize
    \begin{tabular}{@{}p{0.25\textwidth}@{}p{0.75\textwidth}@{}}
    \toprule
    \multirow{4}{*}{Representational level} & Simple concatenation: HVHMC \citep{xu2021hierarchical} \\
    & Custom attention: HBGL \citep{jiang2022exploiting}, PLM-ICD \citep{huang2022plm}, AttentionXML \citep{you2019attentionxml}, LA-HCN \citep{zhang2022hcn}, HiAGM-LA \citep{zhou2020hierarchy} \\
    & Embedding alignment: MatchXML \citep{ye2024matchxml}, PARL \citep{li2024scalable}, HyperIM \citep{chen2020hyperbolic}, ExMLDS \citep{gupta2019distributional} \\
    & Label-informed positive pairs generation: HGCLR \citep{wang2022incorporating} \\
    \midrule
    \multirow{3}{*}{Architectural level} & Graph-based: HiAGM-TP \citep{zhou2020hierarchy}, GalaXC \citep{saini2021galaxc} \\
    & RNN-based: HiAGM-TP \citep{zhou2020hierarchy}, HARNN \citep{huang2019hierarchical}, HILL \citep{zhu2024hill}, HCSM \citep{wang2021cognitive} \\
    & Capsule networks: HCSM \citep{wang2021cognitive} \\
    \midrule
    Parameterization level & Hierarchy-aware parameters sharing or initialization: THMM \citep{pujari2021multi}, HR-SciBERT-mt \citep{sadat2022hierarchical}, HTrans \citep{banerjee2019hierarchical} \\
    \botrule
    \end{tabular}
    \end{minipage}
    
    \vspace{0.5em}
    \noindent\emph{Label info utilization}: Hierarchical label information is used to guide attention mechanisms, embedding alignments, and parameter settings, enhancing the fusion of text and label data.
    
\vspace{1em}

\noindent\begin{minipage}[t]{\textwidth}
    \paragraph{F. Output space segmentation} This submodule re-structures the prediction space, crucial for reducing huge label sets.

    \vspace{0.5em}
    \label{par:output-space}
    \setlength{\tabcolsep}{1pt}
    \footnotesize
    \begin{tabular}{@{}p{0.15\textwidth}@{}p{0.85\textwidth}@{}}
    \toprule
    PLT\textsuperscript{1} & XR-Transformer \citep{zhang2021fast}, MatchXML \citep{ye2024matchxml}, X-Transformer \citep{chang2020taming}, AttentionXML \citep{you2019attentionxml}, Bonsai \citep{khandagale2020bonsai} \\
    \midrule
    ANNS\textsuperscript{2} & GalaXC \citep{saini2021galaxc} \\
    \midrule
    Taxonomy\textsuperscript{3} & THMM \citep{pujari2021multi}, HTrans \citep{banerjee2019hierarchical}, HiLAP \citep{mao2019hierarchical} \\
    \midrule
    Graph-based & ECLARE \citep{mittal2021eclare} \\
    \botrule
    \end{tabular}
    
    \smallskip
    \footnotesize
    \textsuperscript{1}PLT: Probabilistic Label Trees partition the label space into a tree structure for efficient prediction. \\
    \textsuperscript{2}ANNS: Approximate Nearest Neighbor Search uses similarity search to reduce the candidate label space. \\
    \textsuperscript{3}Taxonomy: Direct usage of the given hierarchical structure for label space segmentation.
    \end{minipage}
    
    \vspace{0.5em}
    \noindent\emph{Label info utilization}: Hierarchical label structures are leveraged to segment the output space, improving efficiency and scalability in large label environments.
    
\vspace{1em}

\noindent\begin{minipage}[t]{\textwidth}
    \paragraph{G. Training objectives} This submodule witnesses various loss functions and training strategies, from simple classification losses to sophisticated approaches that capture both semantic relationships and hierarchical structure.

    \vspace{0.5em}

    \label{par:training-obj}
    \setlength{\tabcolsep}{3pt}
    \footnotesize
    \begin{tabular}{@{}p{0.25\textwidth}@{}p{0.75\textwidth}@{}}
    \toprule
    Classification loss & Binary cross entropy: basic building blocks of most methods \\
    \midrule
    \multirow{2}{*}{\parbox{0.25\textwidth}{Multiple\\classification losses}} & Multi-stage: XR-Transformer \citep{zhang2021fast}, MatchXML \citep{ye2024matchxml}, X-Transformer \citep{chang2020taming}, AttentionXML \citep{you2019attentionxml}, Bonsai \citep{khandagale2020bonsai} \\
    & Multi-level: HVHMC \citep{xu2021hierarchical}, HARNN \citep{huang2019hierarchical}, LA-HCN \citep{zhang2022hcn}, HBGL \citep{jiang2022exploiting} \\
    \midrule
    \multirow{2}{*}{Regularization} & Hierarchical constraints: HiAGM \citep{zhou2020hierarchy} \\
    & Label co-occurrence: ExMLDS \citep{gupta2019distributional} \\
    \midrule
    \multirow{3}{*}{Multi-task learning} & Label-prediction pretraining: HBGL \citep{jiang2022exploiting} \\
    & Keyword prediction: HR-SciBERT-mt \citep{sadat2022hierarchical} \\
    & Contrastive learning: XR-Transformer \citep{zhang2021fast}, MatchXML \citep{ye2024matchxml}, X-Transformer \citep{chang2020taming} \\
    & Embedding refinement: MatchXML \citep{ye2024matchxml}, HILL \citep{zhu2024hill}, HGCLR \citep{wang2022incorporating}, ExMLDS \citep{gupta2019distributional}, GalaXC \citep{saini2021galaxc}, ECLARE \citep{mittal2021eclare} \\

    \midrule
    \multirow{3}{*}{Architecture-specific} & Reinforcement learning rewards: HiLAP \citep{mao2019hierarchical} \\
    & Capsule network margin loss: HCSM \citep{wang2021cognitive} \\
    & Ranking loss: PARL \citep{li2024scalable} \\
    \botrule
    \end{tabular}
    \end{minipage}
    
    \vspace{0.5em}
    \noindent\emph{Label info utilization}: Label hierarchical information guides both loss function design and regularization strategies, particularly in multi-stage and multi-task approaches.    
    
\vspace{1em}

\noindent\begin{minipage}[t]{\textwidth}
    \paragraph{H. Score Prediction} This submodule transforms model outputs into label predictions, showing different approaches to the task: basic linear layers treat each label as an independent binary decision, approaches with multiple classifiers each handle labels differently to capture their relationships, while similarity-based methods exploit learned embedding spaces.

    \vspace{0.5em}

    \label{par:score-pred}
    \setlength{\tabcolsep}{3pt}
    \footnotesize
    \begin{tabular}{@{}p{0.25\textwidth}@{}p{0.75\textwidth}@{}}
    \toprule
    Linear & Basic approach: a single linear layer followed by optional sigmoid activation. \\
    \midrule
    \multirow{2}{*}{Multiple Linear} & Multi-stage: XR-Transformer \citep{zhang2021fast}, MatchXML \citep{ye2024matchxml}, X-Transformer \citep{chang2020taming}, AttentionXML \citep{you2019attentionxml} \\
    & Multi-resolution: HVHMC \citep{xu2021hierarchical}, HARNN \citep{huang2019hierarchical}, LA-HCN \citep{zhang2022hcn}, HBGL \citep{jiang2022exploiting} \\
    \midrule
    Similarity-based & HyperIM \citep{chen2020hyperbolic}, ExMLDS \citep{gupta2019distributional}, GalaXC \citep{saini2021galaxc}, ECLARE \citep{mittal2021eclare} \\
    \midrule
    Per-class MLPs & A separate network for each label: THMM \citep{pujari2021multi}, HR-SciBERT-mt \citep{sadat2022hierarchical} \\
    \midrule
    \multirow{3}{*}{Architecture-specific} & Capsule networks: HCSM \citep{wang2021cognitive} \\
    & Policy networks: HiLAP \citep{mao2019hierarchical} \\
    & Learning-to-rank: PARL \citep{li2024scalable} \\
    \botrule
    \end{tabular}

    \end{minipage}
    
    \vspace{0.5em}
    \noindent\emph{Label info utilization}: Label hierarchical information influences prediction strategies through multi-stage architectures, label-specific networks, and specialized prediction mechanisms.
    
\vspace{1em}

\noindent\begin{minipage}[t]{\textwidth}
    \paragraph{I. Prediction refinement} This optional submodule enhances raw predictions through post-processing.

    \vspace{0.5em}
    
    \label{par:pred-refine}
    \setlength{\tabcolsep}{3pt}
    \footnotesize
    \begin{tabular}{@{}p{0.25\textwidth}@{}p{0.75\textwidth}@{}}
    \toprule
    Single-stage ranking & X-Transformer \citep{chang2020taming} \\
    \midrule
    Multi-stage ranking & Hierarchical rankers: XR-Transformer \citep{zhang2021fast}, MatchXML \citep{ye2024matchxml}, X-Transformer \citep{chang2020taming} \\
    \midrule
    Auxiliary networks & After-prediction network to utilize label relationships: CorNet \citep{xun2020correlation} \\
    \midrule
    Post-processing & Data-specific refinement: HCSM \citep{wang2021cognitive} \\
    \botrule
    \end{tabular}
    
    \smallskip
    \end{minipage}
    
    \vspace{0.5em}
    \noindent\emph{Label info utilization}: While CorNet and post-processing rules explicitly leverage label structure, ranking-based approaches focus primarily on computational efficiency for large label spaces rather than hierarchical relationships.
    
\vspace{1em}

Each method discussed in this paper can be characterized by whether and how it instantiates each submodule in the framework, as illustrated chronologically in Figure~\ref{fig:methods-timeline}. While methods may utilize the same types of submodules, their specific choices and implementations within each submodule can vary significantly, with detailed comparisons provided in Table~\ref{tab:submodules}.

\section{Cross-domain Evaluation}\label{sec:evaluation}
In this section, we conduct a comprehensive cross-domain evaluation by selecting representative datasets (Section \ref{sec:datasets}) and methods (Section \ref{sec:selected-methods}) from different domains. We evaluate each method's performance across all datasets using precision and recall metrics (Section \ref{sec:results}), aiming to understand their generalization capabilities beyond their original domains. Section~\ref{sec:observations} briefly discusses the general trends and observations.

\subsection{Datasets}\label{sec:datasets}
We begin by introducing datasets from various domains that are widely used for hierarchical text classification. We present the key characteristics of the datasets and detail noteworthy data cleaning and preprocessing procedures to ensure fair comparisons.

\subsubsection{Selection}\label{sec:data-selection}
We consider datasets from five domains for cross-domain evaluation: legal, scientific, news, medical, and patent classification, as these represent the most common and well-established applications of hierarchical text classification. Our selection criteria focus on datasets with gold-standard taxonomies, textual descriptions, and manageable label spaces (fewer than 4,000 labels), resulting in eight representative datasets: EurLex-3985, EurLex-DC-410, WOS-141, NYT-166, SciHTC-83, SciHTC-800, MIMIC3-3681, and USPTO2M-632. The naming format of the datasets is \texttt{Dataset-N}, where \texttt{N} is the number of labels.

\subsubsection{Characteristics of datasets}\label{sec:data-characteristics}
To facilitate systematic comparison, we characterize each dataset along multiple dimensions: domain, document length statistics, label space properties (cardinality, hierarchy depth), and data distribution metrics (number of labels per sample, number of samples per label). Table~\ref{tab:datasets} presents these detailed statistics.

\begin{table}[h]
    \centering
    \setlength{\tabcolsep}{2pt} %
    \caption{Summary of Datasets}
    \label{tab:datasets}
    \footnotesize %
    \begin{tabular}{llccccccc}
    \toprule
        Dataset & Domain & \makecell{Avg.\\ Len.} & \#Labels & \makecell{Avg/Max/Min \\ \#Labels per sample}& \makecell{Avg/Max/Min \\ \#Samples per label} & \makecell{Max \\ Depth} & \makecell{\#Train\\+Dev} & \#Test \\ 
        \midrule
        EurLex-3956 & Legal & 2635 & 3956 & 5.3/24/1 & 26.0/1253/1 & 2 & 15449 & 3865\\ 
        EurLex-3985* & Legal & 2635 & 3985 & 12.8/38/3 & 62.02/8479/1 & 2 & 15444 & 3862\\ 
        EurLex-DC-410* & Legal & 2635 & 410 & 1.3/7/1 & 61.0/1909/1 & 2 &15472  &  3868\\ 
        WOS-141 & Sci. & 200 & 141 & 2/2/2 & 666.5/14625/1 & 2 & 37588 & 9397 \\ 
        NYT-166 & News & 606 & 166 & 7.6/38/1 & 1665.1/24554/143 & 8 & 29209& 7262\\ 
        SciHTC-83 & Sci. & 145 & 83 & 1.8/2/1 & 4091.2/32854/156 & 6 & 167544& 18616\\ 
        SciHTC-800* & Sci. & 145 & 800 & 1.6/2/1  & 369.8/17166/5 & 6 &167544 & 18616\\ 
        MIMIC3-3681 & Med. & 1514 & 3681 & 15.6/65/1 & 223.2/20046/10 & 0 (+3) & 43978&8734 \\ 
        USPTO2M-632 & Patent & 117 & 632 & 1.3/18/1 & 4239.6/281876/1 & 0 (+2) & 1948508  & 49900 \\ 
        USPTO10k-632** & Patent & 116 & \texttt{"} & 1.89/18/1 & 178.8/10433/1 & \texttt{"}  & 10000  & \texttt{"} \\ 
        USPTO100k-632** & Patent & 116 & \texttt{"} &1.77/18/1 & 418.8/23895/1 & \texttt{"} & 100000  & \texttt{"} \\ \botrule
    \end{tabular}
    \footnotetext{*New variant of the dataset introduced for this study.\\ **Subset created for ablation studies. \texttt{"} indicates same value as USPTO2M-632.\\ (+$n$) in Max Depth indicates that $n$ additional levels were added to the originally flat label annotation to meet the hierarchy requirements for some methods.}
    \end{table} 

\subsubsection{Data preparation}\label{sec:data-preprocessing}
While we maintain consistency with established preprocessing procedures where possible, several datasets required specific handling to ensure compatibility with our evaluation framework, and we also created new versions of some datasets for evaluation. 

\begin{enumerate}
    \item EurLex-3956/3985: Introduced by \cite{loza2008efficient} and known as EurLex-4k \citep{Bhatia16extreme} in the literature, it contains European legal documents sourced from the EUR-Lex repository\footnote{\url{https://eur-lex.europa.eu/homepage.html}}. 
        \begin{enumerate}
            \item Label processing: The original dataset is annotated with 3956 labels of EUROVOC descriptors, describing a wide range of EU-related topics. But the labels are raw text and not mapped to any existing taxonomy. We mapped them to the EUROVOC taxonomy\footnote{\url{https://op.europa.eu/en/web/eu-vocabularies}} using an ensemble of string similarity metrics\footnote{The similarity metrics include simple Levenshtein distance ratio, best partial string matching, Levenshtein distance ratio after sorting words and that ratio between unique sorted word intersections.} and manual verification, resulting in 3890 mapped labels. We then enriched the labels by adding parent codes, yielding a final label set of size 3985. We maintain both versions: EurLex-3956 for literature comparison and EurLex-3985 as our cleaned version for evaluation.
            \item Text processing: We extracted the original text from HTML sources\footnote{\url{https://tudatalib.ulb.tu-darmstadt.de/handle/tudatalib/2937}} rather than using the existing tokenized version from the AttentionXML repository\footnote{\url{https://github.com/yourh/AttentionXML/tree/master\#datasets}} or BOW features from the repository maintained by \cite{Bhatia16extreme}, enabling optimal tokenization for different pretrained language models.
        \end{enumerate}
    \item EurLex-DC-410: Using the original EurLex corpus, we created a new dataset annotated with the Directory Codes (DC). These codes represent classes used in the Directory of Community Legislation in Force. The hierarchy consists of 20 top-level chapter headings with up to four levels of subdivisions. After filtering documents to retain only those with valid DC annotations, the final dataset contains 410 labels.
    \item WOS-141: Introduced by \cite{kowsari2017hdltex}, it contains scientific article abstracts from the Web of Science. The labels represent a hierarchical taxonomy of scientific categories and subcategories, with a total number of 141. We obtained the data from the original source\footnote{\url{https://github.com/kk7nc/HDLTex}}.
    \item NYT-166: Originating from the New York Times Annotated Corpus \citep{sandhaus2008new}, this dataset consists of news articles spanning diverse topics. The 166 labels form a hierarchical taxonomy representing thematic categories from general subjects down to more fine-grained topics.\footnote{The repository of NYT-166 \url{https://catalog.ldc.upenn.edu/LDC2008T19} is not maintained at the time of our experiment, we contacted the authors who have conducted the experiments on this dataset to obtain the data.}
    \item SciHTC-83: Introduced by \cite{sadat2022hierarchical}, this dataset comprises scientific abstracts from multiple research fields. The 83 labels represent a selection of the most frequently occurring subject categories, structured hierarchically to capture broad domains down to more specialized subfields. We use the same processed version provided by \citet{sadat2022hierarchical}.
    \item SciHTC-800: We build SciHTC-800 from the original SciHTC dataset by adding the second-level codes and correcting the inconsistent codes, resulting in a dataset with 800 labels.
    \item MIMIC3-3681: The MIMIC-III clinical database \citep{johnson2016mimic} consists of de-identified hospital discharge summaries annotated with International Classification of Diseases (ICD) codes. These codes span broad medical domains, covering both diagnoses and procedures. We use the MIMIC3-clean variant introduced by \cite{edin2023automated}, which provides a refined version of the dataset optimized for clinical classification tasks.
    \item USPTO2M-632: Introduced by \cite{li2018deeppatent}, this dataset is derived from United States Patent and Trademark Office (USPTO) documents. It contains patent abstracts annotated with hierarchical patent classification codes, reflecting technological fields and subfields. We obtained the data from the original source\footnote{\url{https://github.com/JasonHoou/USPTO-2M}}.
\end{enumerate}

\paragraph{Data adaptation for methods across domains} Since all datasets have given taxonomies, we handle them differently based on each method's requirements:

\begin{enumerate}
    \item For methods that utilize the given taxonomies (e.g., THMM, HILL, HiAGM), we follow their specific preprocessing procedures to incorporate the taxonomies.
    
    \item For methods that don't use the given taxonomies (e.g., XR-Transformer), we simply preprocess the text according to their requirements while ignoring the taxonomies.
\end{enumerate}

Two datasets required special handling due to their flat label annotation structures. MIMIC3-3681 contains only the most specific diagnostic and procedure codes (3681 flat labels), so we expanded its hierarchy by including upper-level codes to create a three-level structure compatible with THMM/HILL/HiAGM. Similarly, USPTO2M-632 contains only subclass labels (632 flat labels), so we added parent and grandparent labels to create a two-level hierarchy. For fair comparison, evaluations on both datasets were still conducted using only the original flat labels.

\subsection{Methods}\label{sec:selected-methods}
We carefully select representative methods for cross-domain evaluation, focusing on those that effectively utilize or can be adapted to use label hierarchies. Our selection process balances comprehensive coverage with practical resource constraints.

\subsubsection{Selection}\label{sec:method-selection}
Resource limitations prevent us from experimenting with all 32 methods surveyed above, therefore we select representative methods based on three criteria:
\begin{enumerate}
    \item Performance: We prioritize methods that achieved state-of-the-art results on their respective datasets, identified through recent surveys (Section~\ref{sec:related}) and 2024 publications.
    \item Label hierarchy utilization: We focus on methods that either integrate label hierarchical information into their main architecture or can be readily adapted to do so, excluding augmentation or post-processing components. Exception is made for PatentBERT/FlatBERT, which serves as a baseline model without label hierarchy information used anywhere.
    \item Implementation adaptability: For reproducibility and extensibility, we prioritize methods with clear architectures that are easy to implement and modify. We exclude ensemble methods to maintain architectural clarity, facilitate potential improvements, and manage computational resources.
\end{enumerate}

\subsubsection{Selected methods for benchmark}\label{sec:benchmark-methods}
Based on these criteria, we selected eight representative methods covering 5 domains:

\emph{Legal Domain}:
The current state-of-the-art on EurLex-3956, \textbf{MatchXML} (2024), extends XR-Transformer by incorporating advanced label encoding mechanisms. Its predecessor, \textbf{XR-Transformer} (2021), remains a widely-cited baseline for hierarchical text classification.\footnote{Originally developed for extreme multi-label classification tasks, MatchXML and XR-Transformer are evaluated on textual datasets with large number of labels, including legal domain, wikipedia, and e-commerce.}

\emph{Scientific and News Domains}:
In these domains, \textbf{HILL} (2024) advances the state-of-the-art on both WOS-141 and NYT-166 through novel hierarchy learning, text-label fusion mechanisms, and contrastive learning objectives. Its predecessor, \textbf{HiAGM-TP} (2020), established the foundation for graph-based label hierarchy modeling. \textbf{HR-SciBERT-mt} (2022) achieves state-of-the-art performance on SciHTC-83, a dataset introduced alongside the model, through multi-task learning and hierarchical classifiers.

\emph{Medical Domain}:
For medical text classification, \textbf{PLM-ICD} (2022) achieves state-of-the-art performance on MIMIC3-3681 by addressing long document handling and incorporating label-aware attention mechanisms.

\emph{Patent Domain}:
In patent classification, PatentBERT/\textbf{FlatBERT} (2020) demonstrates that fine-tuning pretrained language models, such as BERT, with a simple classification head can achieve state-of-the-art performance on USPTO2M-632. \textbf{THMM} (2021) takes a different approach, achieving state-of-the-art results on their own sampled dataset USPTO70k through hierarchical classifier architectures.

Note that some methods face computational and architectural \textbf{constraints}. HiAGM-TP and HILL require complete label hierarchies and substantial memory resources for large label spaces. THMM encounters similar limitations in both hierarchical structure requirements and memory usage. HR-SciBERT-mt presents additional requirements, needing keyword annotations and significant computational resources, particularly for large-scale datasets.

\subsection{Evaluation Metrics}\label{sec:evaluation-metrics}
We evaluate the selected methods using precision@$k$ and recall@$k$, the most widely used metrics for multi-label classification tasks. For a given instance, precision@$k$ is the number of true positive labels in the top-$k$ predictions divided by $k$, while recall@$k$ is the number of true positive labels in the top-$k$  predictions divided by the total number of true labels for that instance. These metrics are then averaged across all test instances. We also report the ranks of the methods on each dataset.

\subsection{Results}\label{sec:results}

We evaluate each method on each dataset. The precision@1 and recall@1 scores are presented in Tables~\ref{tab:cross-domain-precision-scores-ranks} and~\ref{tab:cross-domain-recall-scores-ranks}, respectively, along with the ranks of each method on each dataset. Precision/Recall@3, 5, 8, 10 scores show similar patterns and are presented in Appendix~\ref{sec:additional-results}.  General trends and high-level observations are discussed in the following Section~\ref{sec:observations}, while in-depth analysis is provided later in Section~\ref{sec:discussion}.

All experiments were conducted using PyTorch \citep{paszke2019pytorch} on a single NVIDIA A40 GPU, including our re-implementations of FlatBERT and THMM. We maintained consistent dataset processing and train/test splits across all methods while adhering to each method's specific requirements. For each method-dataset combination, we conducted hyperparameter tuning using a validation set randomly sampled from the training data, with limited search ranges closely following the original papers. All reported results are averages across five random seeds.

To manage computational resources, we established uniform constraints for each experimental run (defined as training one method on one dataset with one random seed using the default epoch count from the original papers). These constraints include a maximum GPU memory usage of 40GB and a time limit of 36 hours. Experiments exceeding these limits are denoted as ``EM'' (exceeds memory) or ``ET'' (exceeds time) in our results.\footnote{We made exceptions for two cases that were completed before establishing these limits: HR-SciBERT-mt on SciHTC-83 and HiAGM-TP on EurLex-3985, both requiring over a week of training time. Their results are included for completeness.}

\emph{Caveat}: While we carefully reproduced all methods following their original implementations and validated against reported results where possible, our evaluation setup may differ from the original papers. For methods reporting the same metrics, our reproduced results closely match published numbers (see \hyperref[par:results-not-included-in-main-comparisons]{the following paragraph} on reproduced results). However, many papers use different evaluation metrics, making direct comparisons impossible. Therefore, the performance numbers reported in this section reflect results under our unified evaluation framework, which enables fair comparisons but may not exactly match previously published results.

\begin{table}[!h]
    \footnotesize
    \centering
    \caption{Cross-Domain Precision@1 Scores and Ranks}
    \label{tab:cross-domain-precision-scores-ranks}
    \begin{tabular}{l@{\hskip 0.1in}cccccccc}
    \toprule
    \textbf{Method} & \makecell{\textbf{WOS}\\-141} & \makecell{\textbf{NYT}\\-166} & \makecell{\textbf{EurLex}\\-3985} & \makecell{\textbf{EurLex-DC}\\-410} & \makecell{\textbf{SciHTC}\\-83} & \makecell{\textbf{SciHTC}\\-800} & \makecell{\textbf{MIMIC3}\\-3681} & \makecell{\textbf{USPTO2M}\\-632} \\ 
    \midrule
    MatchXML      & 86.30 - \textbf{6} & 96.27 - \textbf{2} & 97.26 - \textbf{2} & 89.88 - \textbf{1} & 61.70 - \textbf{1} & 41.28 - \textbf{1} & 86.33 - \textbf{3} & 82.28 - \textbf{1} \\ 
    XR-Transformer & 86.49 - \textbf{5} & 96.64 - \textbf{1} & 97.51 - \textbf{1} & 89.77 - \textbf{2} & 61.53 - \textbf{3} & 40.69 - \textbf{2} & 86.88 - \textbf{2} & 80.99 - \textbf{4} \\ 
    HILL          & 90.30 - \textbf{1} & 95.02 - \textbf{4} & 94.59 - \textbf{4} & 86.12 - \textbf{4} & 57.70 - \textbf{7} & 35.04 - \textbf{7} & 80.12 - \textbf{4} & \texttt{ET} \\ 
    HiAGM-TP      & 89.75 - \textbf{3} & 89.44 - \textbf{7} & 90.76 - \textbf{5} & 85.19 - \textbf{5} & 57.68 - \textbf{8} & 37.57 - \textbf{6} & \texttt{EM} & \texttt{ET} \\ 
    PLM-ICD       & 89.88 - \textbf{2} & 96.03 - \textbf{3} & 94.77 - \textbf{3} & 87.36 - \textbf{3} & 61.40 - \textbf{4} & 38.40 - \textbf{5} & 90.77 - \textbf{1} & 80.73 - \textbf{5} \\ 
    FlatBERT    & 82.65 - \textbf{7} & 90.95 - \textbf{6} & 44.43 - \textbf{7} & 10.08 - \textbf{7} & 60.45 - \textbf{5} & 38.63 - \textbf{4} & 25.84 - \textbf{5} & 81.36 - \textbf{3} \\ 
    THMM          & 89.04 - \textbf{4} & 93.84 - \textbf{5} & 69.94 - \textbf{6} & 49.74 - \textbf{6} & 61.64 - \textbf{2} & 39.17 - \textbf{3} & \texttt{EM} & 81.83 - \textbf{2} \\ 
    HR-SciBERT-mt      & \texttt{ET} & \texttt{ET} & \texttt{ET} & \texttt{ET} & 59.46 - \textbf{6} & \texttt{ET} & \texttt{ET} & \texttt{ET} \\ 
    \botrule
    \end{tabular}
    \footnotetext{\texttt{ET} indicates that the method exceeded the time limit. \texttt{EM} indicates that the method exceeded the memory limit. Same for the following tables.}
\end{table}

\begin{table}[!h]
    \footnotesize
    \centering
    \caption{Cross-Domain Recall@1 Scores and Ranks}
    \label{tab:cross-domain-recall-scores-ranks}
    \begin{tabular}{l@{\hskip 0.1in}cccccccc}
    \toprule
    \textbf{Method} & \makecell{\textbf{WOS}\\-141} & \makecell{\textbf{NYT}\\-166} & \makecell{\textbf{EurLex}\\-3985} & \makecell{\textbf{EurLex-DC}\\-410} & \makecell{\textbf{SciHTC}\\-83} & \makecell{\textbf{SciHTC}\\-800} & \makecell{\textbf{MIMIC3}\\-3681} & \makecell{\textbf{USPTO2M}\\-632} \\ 
    \midrule
    MatchXML      & 43.15 - \textbf{6} & 22.78 - \textbf{2} & 8.06 - \textbf{2} & 78.89 - \textbf{1} & 36.33 - \textbf{2} & 27.84 - \textbf{1} & 7.17 - \textbf{3} & 55.53 - \textbf{1} \\ 
    XR-Transformer & 43.25 - \textbf{5} & 22.93 - \textbf{1} & 8.09 - \textbf{1} & 78.88 - \textbf{2} & 36.29 - \textbf{3} & 27.43 - \textbf{2} & 7.22 - \textbf{2} & 54.75 - \textbf{4} \\ 
    HILL          & 45.15 - \textbf{1} & 22.45 - \textbf{4} & 7.80 - \textbf{4} & 75.64 - \textbf{4} & 33.89 - \textbf{7} & 23.48 - \textbf{7} & 6.59 - \textbf{4} & \texttt{ET} \\ 
    HiAGM-TP      & 44.88 - \textbf{3} & 21.21 - \textbf{7} & 7.49 - \textbf{5} & 74.91 - \textbf{5} & 33.80 - \textbf{8} & 24.77 - \textbf{6} & \texttt{EM} & \texttt{ET} \\
    PLM-ICD       & 44.94 - \textbf{2} & 22.76 - \textbf{3} & 7.84 - \textbf{3} & 76.64 - \textbf{3} & 36.28 - \textbf{4} & 25.24 - \textbf{5} & 7.63 - \textbf{1} & 54.45 - \textbf{5} \\ 
    FlatBERT    & 41.33 - \textbf{7} & 21.45 - \textbf{6} & 3.31 - \textbf{7} & 9.30 - \textbf{7} & 35.78 - \textbf{5} & 25.82 - \textbf{3} & 1.76 - \textbf{5} & 54.92 - \textbf{3} \\ 
    THMM          & 44.52 - \textbf{4} & 22.09 - \textbf{5} & 5.68 - \textbf{6} & 44.77 - \textbf{6} & 36.37 - \textbf{1} & 25.56 - \textbf{4} & \texttt{EM} & 55.26 - \textbf{2} \\ 
    HR-SciBERT-mt      & \texttt{ET} & \texttt{ET} & \texttt{ET} & \texttt{ET} & 34.63 - \textbf{6} & \texttt{ET} & \texttt{ET} & \texttt{ET} \\ 
    \botrule
    \end{tabular}
\end{table}

\paragraph{Results not included in main comparisons}\label{par:results-not-included-in-main-comparisons}
For literature comparison, we validated our implementations against the original EurLex-3956 dataset (also known as EurLex-4k) before using our cleaned version EurLex-3985. Our reproduced precision@1/recall@1 scores (MatchXML: 87.89/17.84, XR-Transformer: 87.83/17.82) closely match reported results (88.12/-, 87.22/- respectively), confirming implementation fidelity.

Analysis of precision and recall scores at higher $k$ values reveals expected precision-recall trade-offs across all datasets: precision decreases while recall increases as $k$ grows. The magnitude of this trade-off varies by dataset: WOS shows dramatic precision drops ($\sim$70 percentage points from $k=1$ to $k=10$), while MIMIC3-3681 exhibits more gradual degradation ($\sim$25 points). Notably, most model rankings remain stable. A more detailed analysis and the numerical results are provided in Appendix~\ref{sec:k-value-analysis}.

\subsection{General observations}\label{sec:observations}
\begin{figure}[!h]
    \centering
    \includegraphics[width=\textwidth]{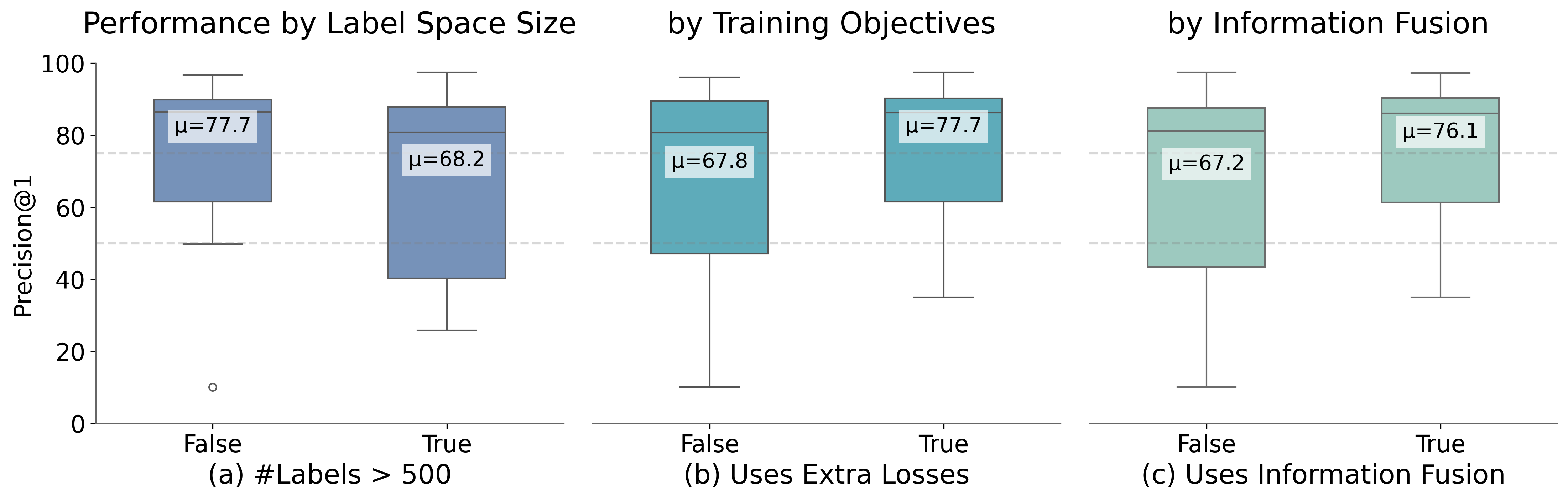}
    \caption{General trends in model performance across different datasets. (a) Label space complexity: Performance decreases with larger label sizses. (b) Model architecture: LLM-based models with advanced learning strategies outperform simpler architectures. (c) Text-label information fusion: Sophisticated mechanisms for combining text and label information yield better results than basic approaches.}
    \label{fig:general-trends}
\end{figure}

Here, we report our high-level observations based on the main cross-domain results, focusing on patterns and insights that emerge from evaluating the methods in their original proposed forms. In the following Section \ref{sec:discussion}, we will conduct more systematic analyses and ablation studies to examine the effects of specific design choices and further dissect the observed trends. Figure~\ref{fig:general-trends} visualizes three main trends:

\emph{Label space complexity} strongly influences model performance, as shown in Fig.~\ref{fig:general-trends}(a). Methods achieve higher scores on datasets with fewer labels (under 500), while performance declines as label count increases. Additionally, as seen from Tables~\ref{tab:datasets} and \ref{tab:cross-domain-recall-scores-ranks}, high per-sample label cardinality (MIMIC3-3681 with 15.6 labels/sample and EurLex-3985 with 12.8 labels/sample) presents challenges, with methods generally achieving lower recall scores on such datasets.

\emph{Model architecture} plays a crucial role, as illustrated in Fig.~\ref{fig:general-trends}(b). Models using large language models as text encoders consistently outperform others, particularly when incorporating advanced learning strategies such as contrastive losses or regularization terms in their training objectives.

\emph{Text-label interaction mechanisms} prove important, demonstrated in Fig.~\ref{fig:general-trends}(c). Methods that effectively combine text and label information through sophisticated mechanisms, such as embedding alignment or label-aware attention, show marked improvements over simpler approaches.

Beyond these salient general trends, some more nuanced patterns emerge across datasets and methods, which are presented in Appendix~\ref{sec:dataset-method-specific-patterns}.

\section{Analysis \& Lessons}\label{sec:discussion}

Our cross-domain evaluation reveals several key insights about hierarchical text classification methods. To better understand these findings, we conduct in-depth analysis and extensive additional experiments investigating: the surprising effectiveness of methods outside their original domains (Section \ref{sec:sota-from-other-domains}), the potential of combining submodules across domains (Section \ref{sec:novel-combinations}), and the impact of various design choices through controlled experiments, including domain-specific language models (Section \ref{sec:impact-llm}), document length handling strategies (Section \ref{sec:impact-long-doc}), training data size variations (Section \ref{sec:impact-data-size}), and label hierarchy initialization methods (Section \ref{sec:impact-tree-init}). Through these systematic analyses and ablation studies, we aim to provide practical guidance for selecting and adapting methods across different domains.

\subsection{State-of-the-art performance often comes from other domains}\label{sec:sota-from-other-domains}
Our cross-domain evaluation reveals interesting patterns in how methods perform beyond their original domains as shown in Table~\ref{tab:sotas-summary}:

\begin{table}[!ht]
    \footnotesize
    \centering
    \caption{Summary of SOTA methods and cross-domain top-performer changes.}
    \label{tab:sotas-summary}
    \begin{tabular}{@{\extracolsep\fill}l@{\hskip 0.15in}l@{\hskip 0.15in}l@{\hskip 0.15in}l@{\hskip 0.15in}l@{\hskip 0.15in}l}
    \toprule
    Dataset & Domain & \multicolumn{1}{c}{SOTA claimed (Status)} & \multicolumn{3}{c}{New Top Performers} \\ 
    & & & \multicolumn{1}{c}{1} & \multicolumn{1}{c}{2} & \multicolumn{1}{c}{3} \\ 
    \midrule
    WOS-141 & Scientific & HILL$^{\text{\sffamily{SN}}}$ (Y) & 
        HILL$^{\text{\sffamily{SN}}}$ & PLM-ICD$^{\text{\sffamily{M}}}$ & HiAGM-TP$^{\text{\sffamily{SN}}}$ \\
    NYT-166 & News & HILL$^{\text{\sffamily{SN}}}$ (\textbf{N}) & 
        XR-Trans$^{\text{\sffamily{L}}}$ & MatchXML$^{\text{\sffamily{L}}}$ & PLM-ICD$^{\text{\sffamily{M}}}$ \\
    EurLex-3985 & Legal & MatchXML$^{\text{\sffamily{L}}}$ (Y) & 
        XR-Trans$^{\text{\sffamily{L}}}$ & MatchXML$^{\text{\sffamily{L}}}$ & PLM-ICD$^{\text{\sffamily{M}}}$ \\
    EurLex-DC-410 & Legal & - & 
        MatchXML$^{\text{\sffamily{L}}}$ & XR-Trans$^{\text{\sffamily{L}}}$ & PLM-ICD$^{\text{\sffamily{M}}}$ \\
    SciHTC-83 & Scientific & HR-SciBERT$^{\text{\sffamily{SN}}}$ (\textbf{N}) & 
        MatchXML$^{\text{\sffamily{L}}}$ & THMM$^{\text{\sffamily{P}}}$ & XR-Trans$^{\text{\sffamily{L}}}$ \\
    SciHTC-800 & Scientific & - & 
        MatchXML$^{\text{\sffamily{L}}}$ & XR-Trans$^{\text{\sffamily{L}}}$ & THMM$^{\text{\sffamily{P}}}$ \\
    MIMIC3-3681 & Medical & PLM-ICD$^{\text{\sffamily{M}}}$ (Y) & 
        PLM-ICD$^{\text{\sffamily{M}}}$ & XR-Trans$^{\text{\sffamily{L}}}$ & MatchXML$^{\text{\sffamily{L}}}$ \\
    USPTO2M-632 & Patent & THMM$^{\text{\sffamily{P}}}$ (\textbf{N}) & 
        MatchXML$^{\text{\sffamily{L}}}$ & THMM$^{\text{\sffamily{P}}}$ & FlatBERT$^{\text{\sffamily{P}}}$ \\
    \botrule
    \end{tabular}
    
    {\raggedright \footnotesize
    Y: SOTA maintained. \\
    \textbf{N: SOTA lost.}\\
    $^{\text{\sffamily{SN}}}$ Methods originally evaluated on Scientific or News domain.  \\
    $^{\text{\sffamily{L}}}$ Methods originally evaluated on Legal domain. \\
    $^{\text{\sffamily{M}}}$ Medical domain methods. \\
    $^{\text{\sffamily{P}}}$ Patent domain methods. 
    \par}
\end{table}

\textbf{WOS-141}: HILL maintains its leading position, but the top three performers now include PLM-ICD, originally designed for medical coding.
\textbf{NYT-166}: Methods from the legal domain, XR-Transformer and MatchXML, have surpassed the previous leader, HILL. PLM-ICD from the medical domain also ranks among the top performers.
\textbf{EurLex-*}: MatchXML and XR-Transformer continue to excel in their native legal domain. Notably, PLM-ICD, from the medical domain, consistently ranks third, indicating that document handling techniques from medical texts can enhance legal document classification.
\textbf{SciHTC-*}: The original leader, HR-SciBERT-mt, has been overtaken by methods from other domains, including MatchXML and XR-Transformer from the legal domain, THMM from the patent domain.
\textbf{MIMIC3-3681}: PLM-ICD retains its top status in the medical domain, while legal domain methods, XR-Transformer and MatchXML, achieve competitive performance.
\textbf{USPTO2M-632}: MatchXML, a method from the legal domain, surpasses the previous leader, THMM.

\paragraph{Hybrid dominance in rankings} The top-3 performers for each dataset all span multiple domains, suggesting that effective hierarchical text classification strategies are often domain-agnostic. Methods originally designed for extreme multi-label classification (MatchXML, XR-Transformer) show particularly strong cross-domain generalization, while medical domain innovations in document handling (PLM-ICD) prove valuable across various technical domains.

\begin{figure}[!ht]
    \centering
    \includegraphics[width=0.65\textwidth]{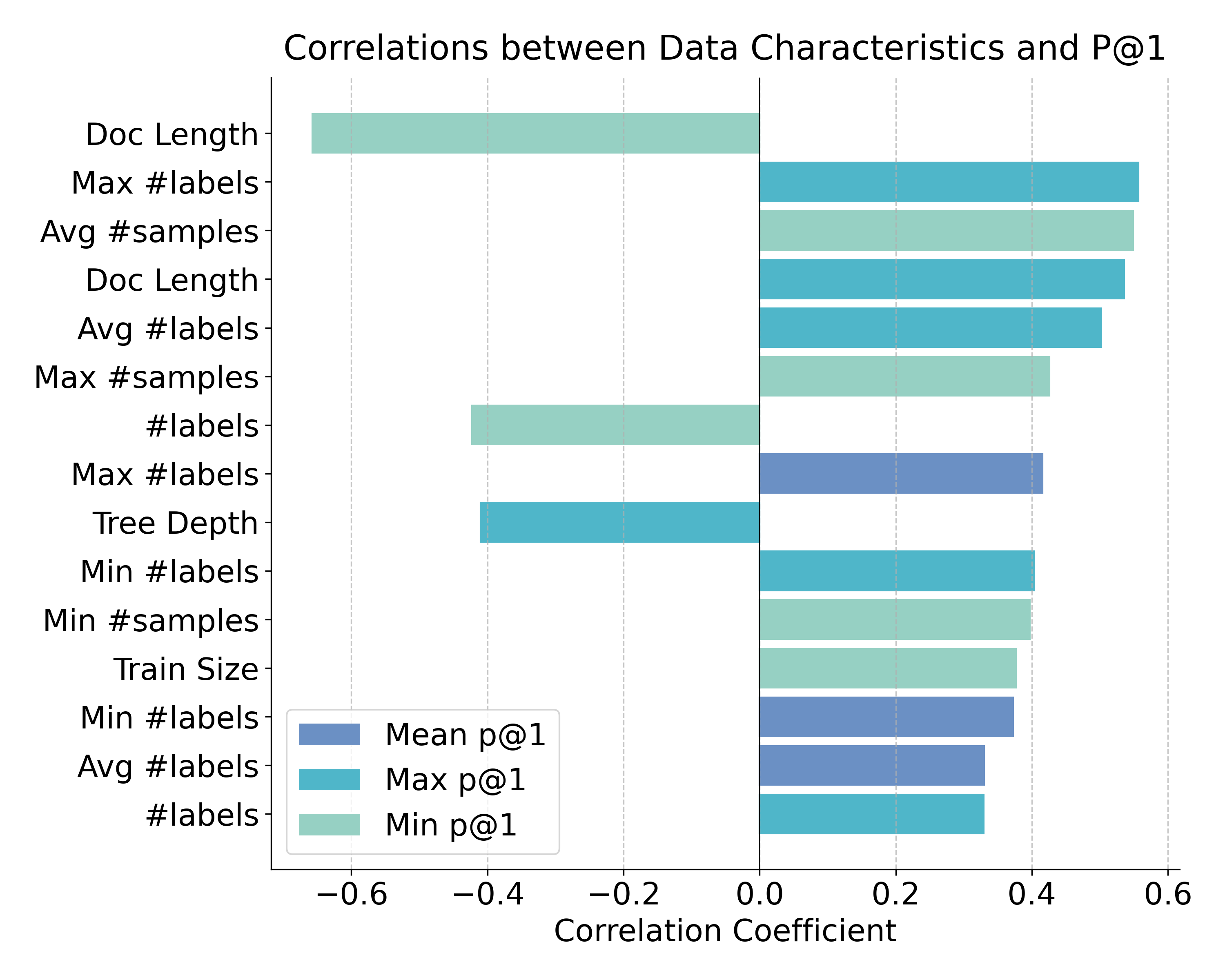}
    \caption{Correlations between dataset characteristics and model performance (precision@1). Only correlations with absolute values greater than 0.3 are shown, with features sorted top-down by absolute correlation values. \\The y-axis shows dataset features, where ``\#labels'' refer to the total number of distinct classes, ``Max/Min/Avg \#labels'' refers to statistics about how many labels each document has, i.e., mean, max, min number of labels per document and similarly ``Max/Min/Avg \#samples'' refers to statistics about how many training examples each label has. Different colored bars show three performance metrics: mean, maximum, and minimum precision@1 scores across all models.}
    \label{fig:dataset_correlation}  
\end{figure}

\paragraph{Dataset characteristics matter more than domain specificity}
To understand what truly drives model performance, we analyzed correlations between dataset features (e.g., document length, labels per sample, training size) and model performance metrics (mean, max, and min precision@1 across models). Figure~\ref{fig:dataset_correlation} shows correlations with absolute values above 0.3.

The results show interesting patterns. \emph{Document length} shows a strong divergent effect ($r = +0.535$ for max precision@1, $r = -0.657$ for min precision@1). Advanced models excel with longer documents, while baseline models perform poorly. Similarly, \emph{label size} shows divergent effect ($r = +0.329$ for max precision@1, $r = -0.423$ for min precision@1).
\emph{Per-sample label cardinality} demonstrates consistent positive impact for above-average models. The average, maximum, and minimum number of labels per sample positively correlate with maximum precision@1 and mean precision@1.
\emph{Label density} significantly benefits weak models. The average, maximum and minimum number of samples per label positively correlate with minimum precision@1 ($r=0.549$ between avg samples per label and min precision@1).
\emph{Hierarchical depth} negatively impacts top performers ($r = -0.410$ for max precision@1).

These findings challenge the common practice of developing and evaluating methods within single domains. It also highlights the need for thorough cross-domain testing and unified evaluation to improve hierarchical text classification.

\subsection{Combining submodules across domains creates new and better models.}\label{sec:novel-combinations}

\begin{figure}[!ht]
    \centering
    \includegraphics[width=0.65\textwidth]{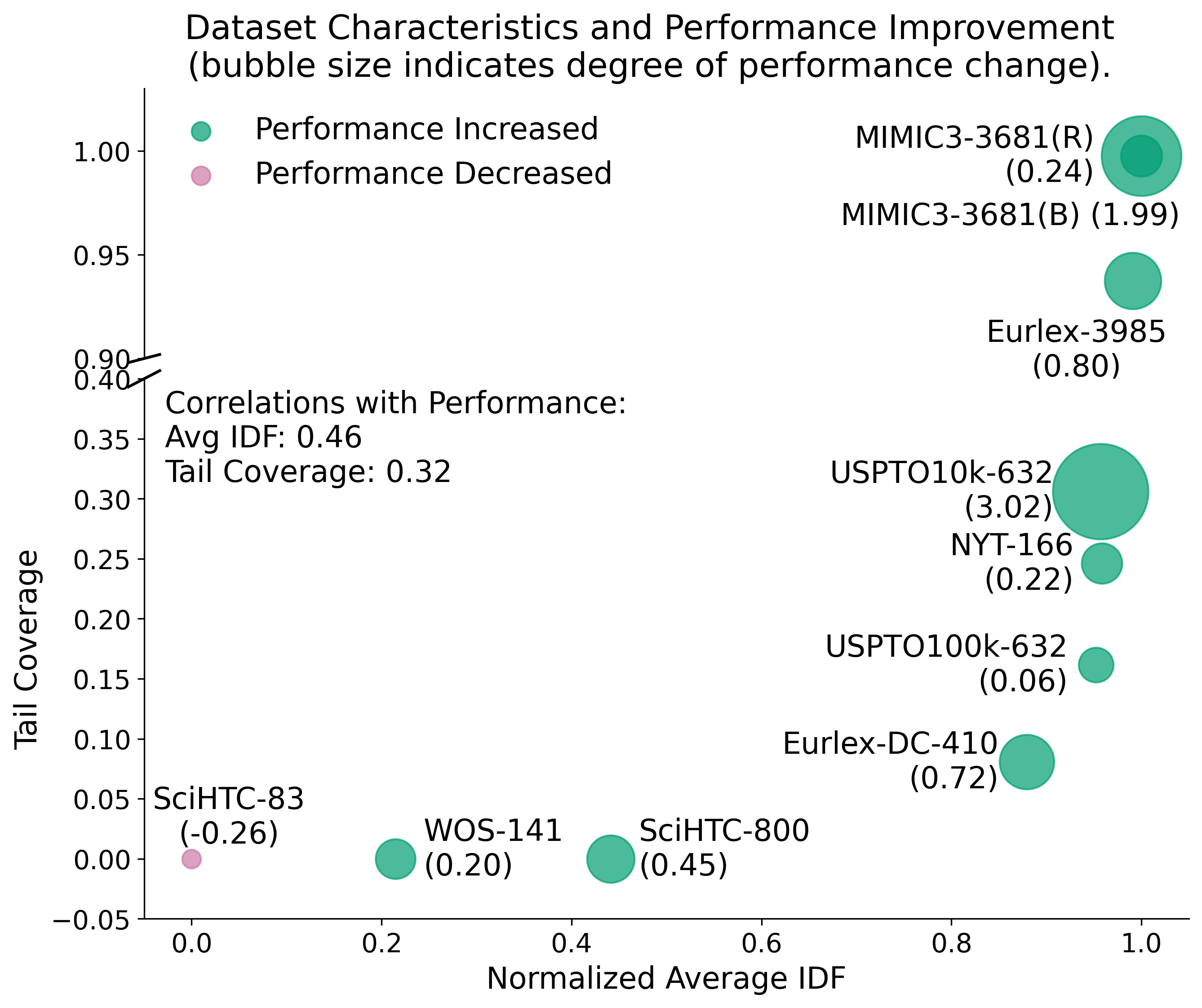}
    \caption{Performance changes from PLM-ICD to PLM-ICD+Label2Vec plotted against dataset characteristics. It shows that augmenting PLM-ICD with label semantic information is beneficial for datasets containing \emph{diverse and rare label combinations}. \\The x-axis shows the average pattern IDF (measuring label combination diversity, see definition~\ref{item:pattern-idf}), and the y-axis shows tail pattern coverage (proportion of samples with rare label combinations, see definition~\ref{item:tail-coverage}). Each point represents a dataset, with larger improvements (shown by point size) occurring in datasets with both high IDF and tail coverage. MIMIC3-3681 results are shown for both BERT (B) and RoBERTa-pm (R) encoders, where RoBERTa-pm is the original text encoder used by PLM-ICD.}
    \label{fig:idf-tail-coverage}
\end{figure}

Both dataset characteristics and limitations of existing methods can guide the combination of submodules across domains to create new and better models. We showcase one such example here.

While PLM-ICD performs strongly across datasets, it lacks label semantic understanding. We hypothesize that incorporating label semantics could improve its performance, particularly for rare label combinations where training data is sparse. To test this, we augment PLM-ICD with Label2Vec (L2V) from MatchXML. The default LLM encoder is bert-base-uncased \citep{devlin2018bert}, and we also test PLM-ICD's originally used RoBERTa-pm \citep{lewis2020pretrained} on MIMIC3-3681.
Table~\ref{tab:plmicd_precision} shows that adding L2V improves performance across most datasets, with exceptions for SciHTC-83. Note that the original PLM-ICD with RoBERTa-pm already performs well on MIMIC3-3681, so the improvement is marginal. Nonetheless, we still observe a new state-of-the-art performance.

We investigate the results by analyzing datasets using the following two metrics based on \emph{label patterns}, i.e. unique sets of labels assigned to individual samples. 
\begin{enumerate}\label{par:label-pattern-metrics}
    \item \label{item:pattern-idf} Average pattern IDF: $\text{IDF}(p) = \frac{1}{|P|}\sum_{p \in P}\log\frac{N}{f_p + 1}$
    where $N$ is number of total samples, $f_p$ is frequency of pattern $p$, and $P$ is the set of unique label patterns. Higher values indicate more diverse label combinations.
    \item \label{item:tail-coverage} Tail pattern coverage: $\frac{\sum_{p: f_p \leq \theta} f_p}{\sum_{p \in P} f_p}$
    where $\theta=3$, measuring the proportion of samples with rare label combinations.
\end{enumerate}

The improvements positively correlate with both measurements. Figure~\ref{fig:idf-tail-coverage} visualizes such correlations. Larger performance gains occur in datasets with both high IDF and tail coverage, suggesting our new method is particularly effective for datasets with many rare label combinations. %

This finding demonstrates how combining architectural innovations from different domains, e.g. PLM-ICD's label-aware attention and long-document handling with MatchXML's label semantics, can create more robust models, especially for challenging scenarios with sparse label patterns.

\begin{table}[t]
    \centering
    \footnotesize
    \caption{Precision@1 comparison of PLM-ICD and PLM-ICD+L2V.}
    \label{tab:plmicd_precision}
    \begin{tabular}{lcccccccccc}
    \toprule
    \textbf{Method} & \makecell{WOS\\-141} & \makecell{NYT\\-166} & \makecell{EurLex\\-3985} & \makecell{EurLex-DC\\-410} & \makecell{SciHTC\\-83} & \makecell{SciHTC\\-800} & \makecell{MIMIC3\\-3681\\BERT} & \makecell{MIMIC3 \\-3681\\RoBERTa-pm} & \makecell{USPTO\\10k\\-632} & \makecell{USPTO\\100k\\-632} \\
    \midrule
    PLM-ICD & 89.88 & 96.03 & 94.77 & 87.36 & 61.40 & 38.40 & 87.13 & 90.77 & 60.74 & 73.21 \\
    +L2V & 90.08 & 96.25 & 95.57 & 88.08 & 61.14 & 38.85 & 89.12 & \textbf{91.01}* & 63.76 & 73.27 \\
    \bottomrule
    \end{tabular}
    \footnotetext{*New state-of-the-art result on MIMIC3-3681.}
\end{table}

\subsection{Domain specific LLMs are beneficial, especially for simpler models and low-resource settings}\label{sec:impact-llm}

\begin{figure}[!h]
    \centering
    \includegraphics[width=\textwidth]{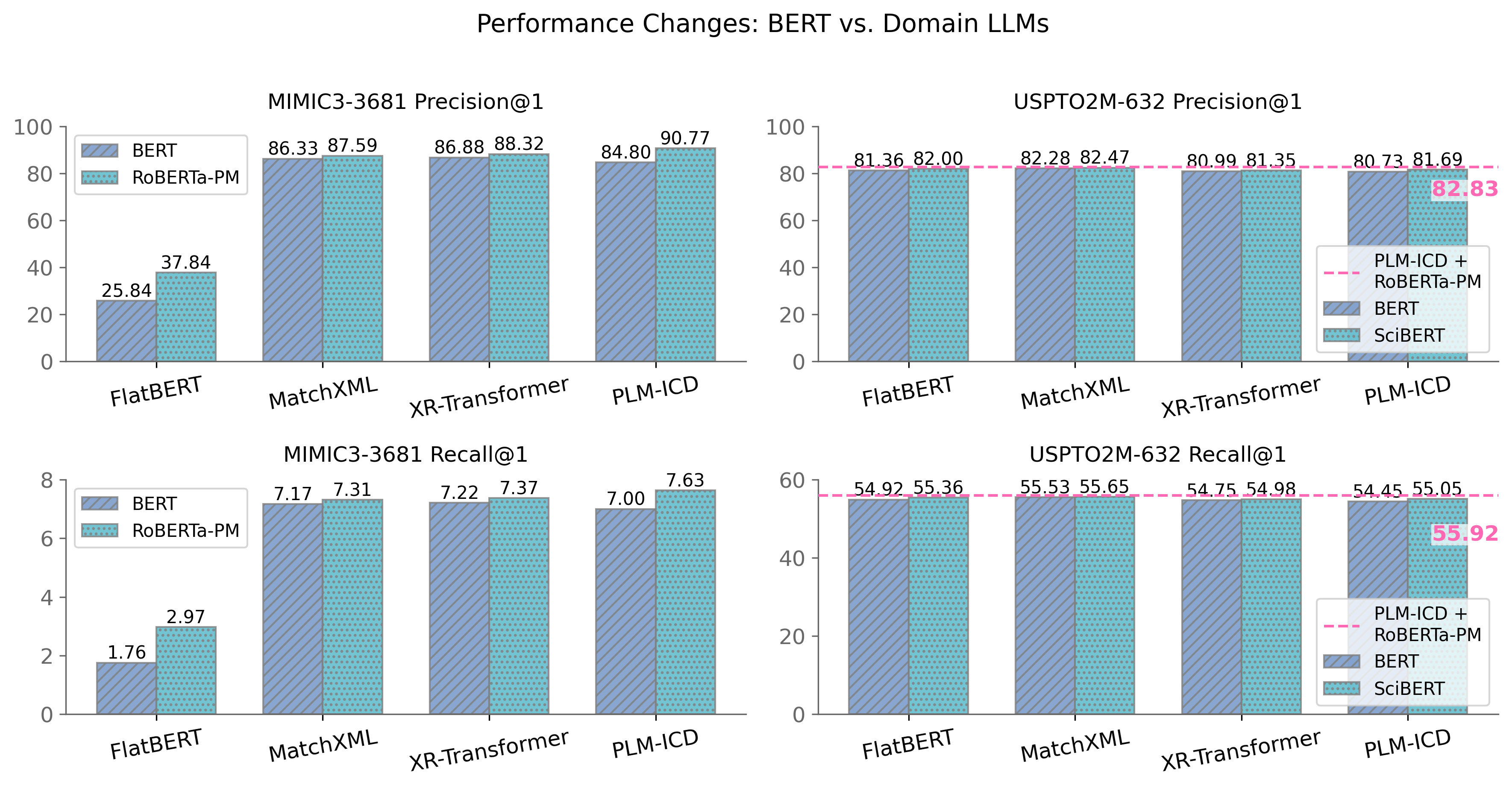}
    \caption{P/R@1 changes from BERT to domain-specific LLMs on MIMIC3-3681 and USPTO2M-632. The paired bars show performance improvements when switching from BERT to domain-specific LLMs. Larger gains are seen on MIMIC3-3681 compared to USPTO2M-632, especially for simpler architectures like FlatBERT. The horizontal dashed line indicates a \emph{new state-of-the-art} achieved by PLM-ICD using RoBERTa-PM (a medical LLM) on USPTO2M-632 (patent), surprisingly outperforming SciBERT (a scientific LLM).}
    \label{fig:impact-llm}
\end{figure}

Among methods using language models as text encoders, most use bert-base-uncased \citep{devlin2018bert} by default, while PLM-ICD and THMM employ domain-specific models (RoBERTa-PM \citep{lewis2020pretrained} and SciBERT \citep{beltagy2019scibert}) on their respective domains, following the original papers. Given PLM-ICD's superior performance on MIMIC3-3681, we investigate whether this advantage comes from its architecture or its domain-specific encoder, as well as the impact of domain-specific LLMs on other methods, with results shown in Figure~\ref{fig:impact-llm}.

All methods using domain-specific LLMs outperform the default bert-base-uncased encoder, and the performance gap is more significant for FlatBERT and MIMIC3-3681.
Two conclusions can be drawn: (1) simpler architectures like FlatBERT benefit more from careful language model selection than complex ones, and (2) the impact of language model choice may diminish with larger training datasets, as evidenced by the smaller performance changes on USPTO2M-632.

PLM-ICD's performance drops significantly when switching from RoBERTa-PM to BERT on MIMIC3-3681, losing its status as state of the art. This indicates that the domain-specific encoder plays a crucial role in its superior performance, while using SciBERT on USPTO2M-632 shows slight improvements. 

\paragraph{A surprising new state of the art with a `mismatched' cross-domain LLM}
In a surprising cross-domain application, using RoBERTa-PM (a medical domain LLM) for PLM-ICD on USPTO2M-632 (a patent dataset) improves performance (80.73/54.45 to 82.83/55.92) even more than using SciBERT (a scientific domain LLM), surpassing the original SOTA from MatchXML. This counter-intuitive success of using a medical LLM for patent classification suggests the improvement may stem from RoBERTa-PM's larger vocabulary rather than domain-specific knowledge. This finding hints at potential benefits of cross-domain LLM applications. Nonetheless, the confirmation of this hypothesis is left as future work.

\subsection{Long document handling is crucial for medical text}\label{sec:impact-long-doc}

Text encoders generally face challenges with long documents, whether due to memory constraints in RNNs or token length limitations in pretrained language models (e.g., 512 tokens for BERT-family models). While most methods simply truncate documents to handle these limitations, our experiments on MIMIC3-3681 reveal the significant impact of long document handling strategies on model performance, shown in Table~\ref{tab:long-doc-impact}.

\begin{table}[h]
    \centering
    \footnotesize
    \caption{Impact of document handling strategies on MIMIC3-3681. PLM-ICD's full document processing outperforms truncation, while naive mean pooling over 512-token segments in XR-Transformer underperforms simple truncation, suggesting effective long document handling requires sophisticated design.}
    \label{tab:long-doc-impact}
    \begin{tabular}{@{\hskip 0.15in}p{0.25\textwidth}@{}p{0.3\textwidth}@{}p{0.4\textwidth}@{}}
    \toprule
    Method &  Change & Performance (P@1/R@1) Change\\
    \midrule
    PLM-ICD & full doc $\rightarrow$ truncated & 90.77/7.63 \textcolor{red}{$\Downarrow$} \textbf{-4.48/-0.46} 86.29/7.17 \\
    XR-Transformer & truncated $\rightarrow$ mean pool & 88.32/7.37 \textcolor{red}{$\Downarrow$} \textbf{-1.94/-0.19} 86.38/7.18 \\
    \bottomrule
    \end{tabular}
\end{table}

PLM-ICD's sophisticated handling of full documents proves beneficial, while our attempt to enhance XR-Transformer beyond its default truncation by averaging embeddings across 512-token segments actually hurts performance. This counter-intuitive result suggests that averaging across document segments may dilute important signals, and effective long document strategies need careful design, potentially incorporating domain knowledge and label-aware mechanisms.

\subsection{Simple models are good enough sometimes, but more easily affected by the data size}\label{sec:impact-data-size}

Our cross-domain evaluation shows that FlatBERT performs poorly on most datasets but achieves competitive results on USPTO2M-632. We hypothesize this exception is due to USPTO2M's large training set (1.9M samples), which compensates for FlatBERT's simple architecture. To test this hypothesis, we created two smaller versions of the dataset: \textbf{USPTO100k-632} with 100k samples and \textbf{USPTO10k-632} with 10k samples, both obtained via stratified sampling while maintaining the original test set. Table~\ref{tab:impact-data-size} shows the results.

\begin{table}[!htb]
    \footnotesize
    \centering
    \caption{Precision/Recall@1 for varied training sizes of USPTO*-632. Simple models like FlatBERT suffer catastrophic drops with reduced data, while sophisticated architectures like XR-Transformer maintain reasonable performance.}
    \label{tab:impact-data-size}
    \begin{tabular*}{\textwidth}{@{\extracolsep{\fill}}l@{\hskip 0.15in}l@{\hskip 0.15in}l@{\hskip 0.15in}l}
    \toprule
    \multicolumn{1}{l}{Method} & \multicolumn{1}{c}{2M} & \multicolumn{1}{c}{100k} & \multicolumn{1}{c}{10k} \\ \midrule
    MatchXML & 82.28/55.53 & 76.23/51.38 \textbf{(-6.05/-4.15)} & 65.05/43.71 \textbf{(-11.18/-7.67)} \\
    XR-Transformer & 80.99/54.75 & 74.94/50.60 \textbf{(-6.05/-4.15)} & 66.34/44.57 \textbf{(-8.60/-6.03)} \\ 
    HILL & \texttt{ET} & 74.59/50.25 & 65.30/43.95 \textbf{(-9.29/-6.30)} \\ 
    HiAGM-TP & \texttt{ET} & 71.19/47.90 & 58.06/39.04 \textbf{(-13.13/-8.86)} \\ 
    PLM-ICD & 80.73/54.45 & 73.21/49.32 \textbf{(-7.52/-5.13)} & 60.74/40.76 \textbf{(-12.47/-8.56)} \\
    FlatBERT & 82.00/55.36 & 69.34/46.54 \textbf{(-12.66/-8.82)} & 17.93/11.78 \textbf{(-51.41/-34.76)} \\ 
    THMM & 81.83/55.26 & 72.00/48.37 \textbf{(-9.83/-6.89)} & 36.98/23.81 \textbf{(-35.02/-24.56)} \\
    \botrule
    \end{tabular*}
\end{table}

The results confirm our hypothesis: FlatBERT's performance drops dramatically with reduced training data, from competitive (82.00/55.36) to catastrophic (17.93/11.78). In contrast, methods with more sophisticated architectures like XR-Transformer maintain reasonable performance even with only 10k samples. This explains FlatBERT's inconsistent performance across datasets: it requires large-scale training data to compensate for its simple architecture.

\subsection{Prior knowledge of label hierarchy for Probabilistic Label Tree is somewhat beneficial}\label{sec:impact-tree-init}

We investigate the impact of different initialization methods for the Probabilistic Label Tree (PLT) in XR-Transformer, which controls how the output space is initially segmented. We compare three approaches: PIFA (the original method using hierarchical clustering based on training samples), Gold (using the ground-truth taxonomy tree), and random initialization. Our experiments show minimal differences between PIFA and Gold, though random initialization leads to decreased performance (see precision and recall details in Tables~\ref{tab:xrt-precision-1} and~\ref{tab:xrt-recall-1}). These results suggest that while prior knowledge of label hierarchies may provide some benefit, the advantage may be modest.

\section{Conclusion}\label{sec:conclusion}
This paper proposes a unified framework for hierarchical multi-label text classification and conducts a comprehensive cross-domain analysis of state-of-the-art methods. Our evaluation reveals several important \textbf{insights}. \emph{First}, top performance often comes from methods developed for other domains, challenging the common practice of domain-specific method development. This is evidenced by our achievement of new state-of-the-art results on NYT-166, SciHTC-83, and USPTO2M-632 using methods from other domains. \emph{Second}, the effectiveness of methods depends more on dataset characteristics (e.g., label patterns, document length, training size) than on the domain it comes from. \emph{Third}, combining architectural innovations from different domains can create more robust models and even new state-of-the-art results, as demonstrated by our experiments in Sec.~\ref{sec:novel-combinations}.

Other key \textbf{findings} include: domain-specific language models significantly improve performance, especially for simpler architectures and challenging datasets; sophisticated document handling is crucial for long texts like medical records; simple architectures can achieve competitive performance but require substantial training data; and prior knowledge of label hierarchies provides modest benefits.

Our study has several \textbf{limitations} that highlight opportunities for future research. On the resource side, computational constraints limited our evaluation to a subset of available methods and datasets, potentially missing valuable insights from other approaches. This also affected our hyperparameter optimization, possibly underestimating some methods' full potential. Our unified evaluation framework, while enabling fair comparisons across methods, may yield results that differ from those reported in original papers due to standardized preprocessing and evaluation settings. Regarding methodological scope, our focus on end-to-end architectures excluded standalone components like data augmentation or post-processing methods that could provide complementary benefits. Additionally, methods for semi-supervised, unsupervised, or zero-shot approaches remain to be explored.

Several promising research \textbf{directions} emerge from our findings. \emph{First}, the success of combining PLM-ICD with Label2Vec suggests potential in systematically exploring other cross-domain architectural combinations. 
\emph{Second}, the significant impact of document handling strategies on medical texts indicates a need for more sophisticated approaches that can effectively process long documents while maintaining computational efficiency. 
\emph{Third}, developing methods that can maintain performance with limited training data or computational resources would increase practical applicability across domains. 
\emph{Fourth}, recent advances in large language models suggest opportunities for effective zero-shot and few-shot classification with minimal domain-specific training data. 
\emph{Lastly}, extending evaluation to emerging domains like skill tagging \citep{li2023skillgpt, li2023llm4jobs} could reveal additional insights about method generalization and domain-specific challenges.

Looking ahead, the key take-away of our study is that it is valuable for researchers and practitioners to consider possibilities beyond domain boundaries in hierarchical text classification. Rather than defaulting to domain-specific solutions, we should explore methods that have succeeded elsewhere. This shift in approach could accelerate progress and lead to more robust and effective classification systems across all domains.

\section*{Acknowledgments}
The research leading to these results has received funding from the Special Research Fund (BOF) of Ghent University (BOF20/IBF/117), from the Flemish Government under the ``Onderzoeksprogramma Artificiële Intelligentie (AI) Vlaanderen'' programme, from the FWO (project no. G0F9816N, 3G042220, G073924N). Funded by the European Union (ERC, VIGILIA, 101142229). Views and opinions expressed are however those of the author(s) only and do not necessarily reflect those of the European Union or the European Research Council Executive Agency. Neither the European Union nor the granting authority can be held responsible for them.

\bibliography{ref}%

\newpage

\begin{appendices}
\section{Details of the submodules used in recent methods}\label{sec:details-submodules}

Table~\ref{tab:submodules} shows the submodules used in recent methods.

\begin{sidewaystable}[!h]
    \tiny
    \centering
    \caption{Submodules used in recent methods}
    \label{tab:submodules}
    \begin{tabular}{cccccccc}
    \toprule
    Method & Text Encoder & Label Encoder & Out space seg & Text-Label info fusion & Training objectives & Score Prediction & Prediction Refinement \\ 
    \midrule
    XR-Transformer & LLM\footnotemark[0]\&TF-IDF & PIFA & PLT & - & \makecell{(Multi-stage) CLS loss\footnotemark[1]+ \\Contrastive (MAN+TFN)} & MS Linear & MS ranker \\ 
    MatchXML & LLM\footnotemark[0]\&TF-IDF & Label2Vec & PLT & Embedding alignment & \makecell{(Multi-stage) CLS loss + \\ Contrastive (MAN+TFN) \\+ Alignment loss} & MS Linear &  MS ranker \\ 
    HILL & LLM\footnotemark[0] & Coding Tree & - & Structure encoder\footnotemark[2] & \makecell{CLS loss + \\NT-Xent contrastive loss}& Linear & -\\ 
    HiAGM-TP & GRU + CNN & Tree-LSTM/GCN & - & Structure encoder\footnotemark[3] & CLS loss with regularization& Linear& -\\ 
    HiAGM-LA & GRU + CNN & Tree-LSTM/GCN & - & Structured label attn & CLS loss with regularization& Linear& -\\ 
    PLM-ICD & LLM\footnotemark[0] & -& -& \makecell{Segmented text\\ with label-aware attn}& CLS loss & Linear & - \\ 
    FlatBERT\footnotemark[4] & LLM\footnotemark[0] & - & - & - & CLS loss & Linear & -\\ 
    THMM & LLM\footnotemark[0] & - & Original taxonomy & \makecell{Parent-kid \\representation concatenation} & CLS loss & An MLP per class & - \\ 
    HR-SciBERT-mt & LLM\footnotemark[0] & - & Original taxonomy & \makecell{Parent-kid \\parameter sharing} & \makecell{CLS loss + \\Keyword prediction loss}  & An MLP per class & - \\ 
    X-Transformer & LLM \& TF-IDF &  PIFA  & PLT & - &  \makecell{CLS loss + \\Contrastive (MAN+TFN)} & Linear & 1-stage ranker \\ 
    XR-Linear & TF-IDF & PIFA & PLT & - &  \makecell{MS CLS loss + \\Contrastive (MAN+TFN)} & MS Linear & MS ranker  \\ 
    AttentionXML & LSTM & PIFA & PLT & label attn & (Multi-Stage) CLS & MS Linear & -  \\ 
    Bonsai & TF-IDF & PIFA & PLT  & - & (Multi-Stage) CLS & MS Linear &  - \\ 
    HVHMC & GCN (word-doc)  & \makecell{GCN(label-word) \\ inter+intra level dependency} & - & repr concat & \makecell{local + \\global CLS} & \makecell{level-wise +\\all cat} & - \\ 
    HGCLR & LLM & Graphnomer & - & \makecell{label guided \\ positive sample gen} & \makecell{CLS +\\ NT-Xent contrastive} & Linear & - \\ 
    HBGL & LLM & LLM with custom pretraining task & - & \makecell{Repr concat +\\ custom attn mask} & \makecell{Multi-stage CLS +\\ label mask pretraining} & level-wise & - \\ 
    HARNN & LSTM & dense vec & - & HAM &  \makecell{local + \\global CLS} & \makecell{level-wise +\\all cat} & -  \\ 
    HTrans & \makecell{GRU +\\modified outputs} & - & Original taxonomy & \makecell{Parent-kid \\parameter sharing} & CLS &  An MLP per class & -  \\ 
    PARL & \makecell{LLM (bi+cross) +\\GraphSAGE+\\BM25} & LLM & - & Bi-enc + Cross-enc & Ranknet loss &  L2R & - \\  
    HyperIM & Hyperbolic & Hyperbolic & - & \makecell{Label-aware\\doc repr\\(geodesic distance)} & \makecell{CLS loss + \\Contrastive} & emb sim & - \\ 
    LA-HCN & LSTM & dense vec & - & \makecell{Level-wise label attn+\\label-component assoc.} & \makecell{local + \\global CLS} & \makecell{combine\\local+global} & - \\ 
    HCSM & CNN+LSTM & w2v  & - & \makecell{HARNN+\\HON-LSTM+\\HBiCaps} & Margin loss & 1 cap for 1 class & - \\ 
    HiLAP & CNN & dense vec & Original taxonomy & - & \makecell{Neg discounted\\cum rewards} & Policy network & - \\ 
    ExMLDS & \makecell{TF-IDF+\\ADMM projection} & \makecell{skip-gram +\\SPPMI factorization} & - & ADMM projection & \makecell{CLS +\\skig-gram neg+\\label co-occurance reg} & emb similarity & - \\ 
    GalaXC & Fixed pre-trained & \makecell{GNN +\\label attention} & ANNS & Joint doc-label graph  & \makecell{CLS loss + \\neg Contrastive} & emb sim & - \\ 
    EcLARE & TF-IDF & \makecell{GCN + \\label attention} & \makecell{Graph-Assisted\\Multi-label\\Expansion (GAME)}  & - & \makecell{CLS loss + \\neg Contrastive +\\spectral reg} & emb sim & - \\ 
    \midrule
    \multicolumn{8}{l}{\textbf{Plug-in Methods (can be combined with methods above):}} \\
    \multicolumn{8}{l}{\textit{Training Objective Enhancements:}} \\
    DepReg & \multicolumn{7}{l}{Adds dependency regularization term to training objectives} \\
    \multicolumn{8}{l}{\textit{Label Space Processing:}} \\
    PS-PLT & \multicolumn{7}{l}{Enhances PLT with propensity scoring for better label space segmentation} \\
    \multicolumn{8}{l}{\textit{Prediction Refinement:}} \\
    Capsule Net & \multicolumn{7}{l}{Replaces standard prediction with capsule network + label correlation restrictions} \\
    CorNet & \multicolumn{7}{l}{Adds correlation-based prediction refinement layer} \\
    \multicolumn{8}{l}{\textit{Data Augmentation:}} \\
    GANDALF & \multicolumn{7}{l}{Graph-based data augmentation for long-tail labels} \\
    REMIDIAL & \multicolumn{7}{l}{Data augmentation through label correlation patterns} \\
    \botrule
    \end{tabular}
        \footnotetext[0]{LLM: Large Language Model, e.g. BERT, RoBERTa, etc. Specifically, we use BERT as the default LLM encoder except for the following cases: RoBERTa-PM for PLM-ICD on MIMIC3-3681 and SciBERT for THMM on USPTO2M-632 following the original papers.}
        \footnotetext[1]{Classification loss for multi-label classification, depending on the specific loss function used, most commonly the binary cross-entropy loss.}
        \footnotetext[2]{A custom defined structure encoder that propagates information from the leaf nodes to the root node of the coding tree.}
        \footnotetext[3]{Tree-LSTM or GCN, corresponding to the label encoder.}
        \footnotetext[4]{PatentBERT in the original paper due to its application in patent classification. We rename it as FlatBERT for generalization.}
\end{sidewaystable}

\section{Additional experiment results}\label{sec:additional-results}
\subsection{Impact of Different k Values}\label{sec:k-value-analysis}
While the main text focuses on Precision/Recall@1, analyzing model behaviors at k=3, 5, 8, 10 reveals additional insights into multi-label prediction capabilities (Tables~\ref{tab:cross-domain-precision-scores-ranks-3}, \ref{tab:cross-domain-recall-scores-ranks-3}, \ref{tab:cross-domain-precision-scores-ranks-5}, \ref{tab:cross-domain-recall-scores-ranks-5}, \ref{tab:cross-domain-precision-scores-ranks-8}, \ref{tab:cross-domain-recall-scores-ranks-8}, \ref{tab:cross-domain-precision-scores-ranks-10}, \ref{tab:cross-domain-recall-scores-ranks-10})

\paragraph{Precision-Recall Trade-offs} All datasets show precision decreases and recall increases as $k$ grows, but with notably different magnitudes. WOS shows dramatic precision drops ($\sim$70 percentage points from $k=1$ to $k=10$, e.g., HILL: 90.81$\rightarrow$19.31) due to its having only two labels per document. MIMIC3 exhibits more gradual degradation ($\sim$25 points, e.g., PLM-ICD: 90.77$\rightarrow$67.11) for its high label cardinality. EurLex-3985 demonstrates the most stable precision metrics with only $\sim$6-7 point drops from $k=1$ to $k=5$ (XR-Transformer: 97.67$\rightarrow$90.91).

\paragraph{Model Ranking Changes}
Analysis of model rankings across $k=1,3,5,8,10$ reveals several consistent patterns and notable shifts. Some models maintain stable leadership positions: XR-Transformer consistently ranks first on NYT-166 and EurLex-3985, HILL maintains first place on WOS-141, and PLM-ICD leads on MIMIC3-3681 across all $k$ values. However, significant ranking changes emerge on certain datasets. PLM-ICD shows marked improvement on SciHTC-83, climbing from rank 4 (k=1) to rank 1 (k=8,10), while HILL demonstrates similar improvement on EurLexDC-410, advancing from rank 4 to rank 1. Conversely, some models experience performance degradation at higher k values: MatchXML drops from rank 1 to rank 5 on SciHTC-83, and XR-Transformer declines from rank 3 to rank 7 on the same dataset. The most stable rankings across k values are observed for XR-Transformer on NYT-166, EurLex-3985, and USPTO2M-632, HILL on WOS-141, and PLM-ICD on MIMIC3-3681.

\begin{table}[!h]
    \footnotesize
    \centering
    \caption{Cross-Domain Precision@3 Scores and Ranks}
    \label{tab:cross-domain-precision-scores-ranks-3}
    \begin{tabular}{l@{\hskip 0.15in}cccccccc}
    \toprule
    \textbf{Method} & \makecell{\textbf{WOS} \\-141} & \makecell{\textbf{NYT} \\-166} & \makecell{\textbf{EurLex} \\-3985} & \makecell{\textbf{EurLexDC} \\-410} & \makecell{\textbf{SciHTC} \\-83} & \makecell{\textbf{SciHTC} \\-800} & \makecell{\textbf{MIMIC3} \\-3681} & \makecell{\textbf{USPTO2M} \\-632} \\
    \midrule
    MatchXML & 57.28 - 5 & 84.67 - 2 & 94.36 - 2 & 38.92 - 2 & 39.90 - 1 & 25.79 - 1 & 79.42 - 3 & 44.22 - 3 \\
    XR-Transformer & 57.65 - 4 & 85.16 - 1 & 94.67 - 1 & 39.02 - 1 & 39.35 - 5 & 25.15 - 2 & 80.04 - 2 & 42.75 - 4 \\
    HILL & 60.23 - 1 & 81.83 - 4 & 92.27 - 3 & 37.94 - 3 & 36.56 - 8 & 21.84 - 7 & 71.31 - 4 & \texttt{ET} \\
    HiAGM-TP & 60.10 - 2 & 73.76 - 7 & 87.19 - 5 & 37.19 - 5 & 36.93 - 7 & 23.35 - 6 & \texttt{ET} & \texttt{ET} \\
    PLM-ICD & 59.17 - 3 & 84.04 - 3 & 91.37 - 4 & 37.81 - 4 & 39.65 - 3 & 24.04 - 4 & 85.64 - 1 & 44.57 - 2 \\
    FlatBERT & 34.81 - 7 & 75.23 - 6 & 42.53 - 7 & 6.55 - 7 & 38.99 - 6 & 23.95 - 5 & 19.46 - 5 & 44.85 - 1 \\
    THMM & 57.24 - 6 & 81.74 - 5 & 54.32 - 6 & 22.17 - 6 & 39.83 - 2 & 24.63 - 3 & \texttt{ET} & \texttt{ET} \\
    HR-SciBERT-mt & \texttt{ET} & \texttt{ET} & \texttt{ET} & \texttt{ET} & 39.44 - 4 & \texttt{ET} & \texttt{ET} & \texttt{ET} \\
    \botrule
    \end{tabular}
    \end{table}

    \begin{table}[!h]
    \footnotesize
    \centering
    \caption{Cross-Domain Recall@3 Scores and Ranks}
    \label{tab:cross-domain-recall-scores-ranks-3}
    \begin{tabular}{l@{\hskip 0.15in}cccccccc}
    \toprule
    \textbf{Method} & \makecell{\textbf{WOS} \\-141} & \makecell{\textbf{NYT} \\-166} & \makecell{\textbf{EurLex} \\-3985} & \makecell{\textbf{EurLexDC} \\-410} & \makecell{\textbf{SciHTC} \\-83} & \makecell{\textbf{SciHTC} \\-800} & \makecell{\textbf{MIMIC3} \\-3681} & \makecell{\textbf{USPTO2M} \\-632} \\
    \midrule
    MatchXML & 85.93 - 5 & 53.97 - 2 & 23.36 - 2 & 93.28 - 2 & 66.70 - 1 & 50.30 - 1 & 19.22 - 3 & 78.23 - 3 \\
    XR-Transformer & 86.47 - 4 & 54.36 - 1 & 23.46 - 1 & 93.62 - 1 & 65.83 - 4 & 48.94 - 2 & 19.38 - 2 & 76.19 - 4 \\
    HILL & 90.34 - 1 & 51.33 - 5 & 22.79 - 3 & 91.62 - 3 & 60.98 - 8 & 42.26 - 7 & 17.01 - 4 & \texttt{ET} \\
    HiAGM-TP & 90.16 - 2 & 47.09 - 6 & 21.49 - 5 & 89.89 - 5 & 61.41 - 7 & 45.05 - 6 & \texttt{ET} & \texttt{ET} \\
    PLM-ICD & 88.75 - 3 & 53.45 - 3 & 22.60 - 4 & 90.96 - 4 & 66.34 - 3 & 46.44 - 5 & 20.94 - 1 & 78.70 - 2 \\
    FlatBERT & 52.22 - 7 & 46.08 - 7 & 9.87 - 7 & 18.40 - 7 & 65.41 - 6 & 47.27 - 4 & 4.15 - 5 & 79.04 - 1 \\
    THMM & 85.86 - 6 & 51.42 - 4 & 12.82 - 6 & 56.37 - 6 & 66.54 - 2 & 47.59 - 3 & \texttt{ET} & \texttt{ET} \\
    HR-SciBERT-mt & \texttt{ET} & \texttt{ET} & \texttt{ET} & \texttt{ET} & 65.78 - 5 & \texttt{ET} & \texttt{ET} & \texttt{ET} \\
    \botrule
    \end{tabular}
    \end{table}

    \begin{table}[!h]
    \footnotesize
    \centering
    \caption{Cross-Domain Precision@5 Scores and Ranks}
    \label{tab:cross-domain-precision-scores-ranks-5}
    \begin{tabular}{l@{\hskip 0.15in}cccccccc}
    \toprule
    \textbf{Method} & \makecell{\textbf{WOS} \\-141} & \makecell{\textbf{NYT} \\-166} & \makecell{\textbf{EurLex} \\-3985} & \makecell{\textbf{EurLexDC} \\-410} & \makecell{\textbf{SciHTC} \\-83} & \makecell{\textbf{SciHTC} \\-800} & \makecell{\textbf{MIMIC3} \\-3681} & \makecell{\textbf{USPTO2M} \\-632} \\
    \midrule
    MatchXML & 36.29 - 5 & 72.24 - 2 & 90.69 - 2 & 23.87 - 2 & 27.73 - 4 & 18.52 - 1 & 72.77 - 3 & 29.75 - 3 \\
    XR-Transformer & 36.67 - 4 & 72.58 - 1 & 90.91 - 1 & 24.01 - 1 & 27.30 - 6 & 17.93 - 3 & 73.32 - 2 & 28.60 - 4 \\
    HILL & 37.53 - 1 & 69.69 - 5 & 88.94 - 3 & 23.70 - 3 & 25.75 - 8 & 15.97 - 7 & 64.29 - 4 & \texttt{ET} \\
    HiAGM-TP & 37.39 - 2 & 61.81 - 7 & 82.97 - 5 & 23.22 - 5 & 26.47 - 7 & 17.11 - 6 & \texttt{ET} & \texttt{ET} \\
    PLM-ICD & 37.10 - 3 & 71.89 - 3 & 87.11 - 4 & 23.40 - 4 & 27.97 - 2 & 17.80 - 4 & 80.02 - 1 & 30.46 - 2 \\
    FlatBERT & 23.77 - 7 & 64.12 - 6 & 37.70 - 7 & 4.15 - 7 & 27.56 - 5 & 17.52 - 5 & 14.38 - 5 & 30.71 - 1 \\
    THMM & 35.73 - 6 & 69.73 - 4 & 47.55 - 6 & 14.79 - 6 & 28.05 - 1 & 18.05 - 2 & \texttt{ET} & \texttt{ET} \\
    HR-SciBERT-mt & \texttt{ET} & \texttt{ET} & \texttt{ET} & \texttt{ET} & 27.95 - 3 & \texttt{ET} & \texttt{ET} & \texttt{ET} \\
    \botrule
    \end{tabular}
    \end{table}

    \begin{table}[!h]
    \footnotesize
    \centering
    \caption{Cross-Domain Recall@5 Scores and Ranks}
    \label{tab:cross-domain-recall-scores-ranks-5}
    \begin{tabular}{l@{\hskip 0.15in}cccccccc}
    \toprule
    \textbf{Method} & \makecell{\textbf{WOS} \\-141} & \makecell{\textbf{NYT} \\-166} & \makecell{\textbf{EurLex} \\-3985} & \makecell{\textbf{EurLexDC} \\-410} & \makecell{\textbf{SciHTC} \\-83} & \makecell{\textbf{SciHTC} \\-800} & \makecell{\textbf{MIMIC3} \\-3681} & \makecell{\textbf{USPTO2M} \\-632} \\
    \midrule
    MatchXML & 90.72 - 5 & 66.57 - 2 & 37.20 - 2 & 94.74 - 2 & 76.71 - 4 & 59.79 - 1 & 28.58 - 3 & 84.44 - 3 \\
    XR-Transformer & 91.67 - 4 & 66.88 - 1 & 37.31 - 1 & 95.31 - 1 & 75.53 - 6 & 57.67 - 3 & 28.80 - 2 & 81.97 - 4 \\
    HILL & 93.83 - 1 & 63.00 - 5 & 36.39 - 3 & 94.20 - 3 & 71.33 - 8 & 51.54 - 7 & 24.94 - 4 & \texttt{ET} \\
    HiAGM-TP & 93.48 - 2 & 57.34 - 6 & 33.81 - 5 & 92.76 - 5 & 73.01 - 7 & 55.22 - 6 & \texttt{ET} & \texttt{ET} \\
    PLM-ICD & 92.75 - 3 & 66.04 - 3 & 35.67 - 4 & 93.08 - 4 & 77.41 - 2 & 57.43 - 4 & 31.67 - 1 & 85.93 - 2 \\
    FlatBERT & 59.41 - 7 & 56.46 - 7 & 14.53 - 7 & 18.99 - 7 & 76.40 - 5 & 57.35 - 5 & 5.19 - 5 & 86.40 - 1 \\
    THMM & 89.33 - 6 & 63.77 - 4 & 18.61 - 6 & 61.25 - 6 & 77.58 - 1 & 58.17 - 2 & \texttt{ET} & \texttt{ET} \\
    HR-SciBERT-mt & \texttt{ET} & \texttt{ET} & \texttt{ET} & \texttt{ET} & 77.21 - 3 & \texttt{ET} & \texttt{ET} & \texttt{ET} \\
    \botrule
    \end{tabular}
    \end{table}

    \begin{table}[!h]
    \footnotesize
    \centering
    \caption{Cross-Domain Precision@8 Scores and Ranks}
    \label{tab:cross-domain-precision-scores-ranks-8}
    \begin{tabular}{l@{\hskip 0.15in}cccccccc}
    \toprule
    \textbf{Method} & \makecell{\textbf{WOS} \\-141} & \makecell{\textbf{NYT} \\-166} & \makecell{\textbf{EurLex} \\-3985} & \makecell{\textbf{EurLexDC} \\-410} & \makecell{\textbf{SciHTC} \\-83} & \makecell{\textbf{SciHTC} \\-800} & \makecell{\textbf{MIMIC3} \\-3681} & \makecell{\textbf{USPTO2M} \\-632} \\
    \midrule
    MatchXML & 23.36 - 5 & 60.60 - 2 & 83.40 - 2 & 15.07 - 3 & 19.03 - 5 & 13.10 - 1 & 63.90 - 3 & 19.83 - 3 \\
    XR-Transformer & 23.56 - 4 & 61.04 - 1 & 83.55 - 1 & 15.16 - 2 & 18.67 - 6 & 12.63 - 4 & 64.22 - 2 & 19.02 - 4 \\
    HILL & 23.96 - 1 & 58.32 - 4 & 82.18 - 3 & 15.19 - 1 & 17.91 - 8 & 11.56 - 7 & 55.18 - 4 & \texttt{ET} \\
    HiAGM-TP & 23.82 - 2 & 50.89 - 7 & 75.94 - 5 & 14.84 - 5 & 18.64 - 7 & 12.43 - 6 & \texttt{ET} & \texttt{ET} \\
    PLM-ICD & 23.74 - 3 & 60.53 - 3 & 79.29 - 4 & 14.90 - 4 & 19.52 - 1 & 12.95 - 3 & 72.07 - 1 & 20.60 - 2 \\
    FlatBERT & 16.84 - 7 & 54.00 - 6 & 31.73 - 7 & 2.63 - 7 & 19.19 - 4 & 12.61 - 5 & 8.99 - 5 & 20.79 - 1 \\
    THMM & 23.02 - 6 & 58.14 - 5 & 39.24 - 6 & 10.21 - 6 & 19.43 - 3 & 13.04 - 2 & \texttt{ET} & \texttt{ET} \\
    HR-SciBERT-mt & \texttt{ET} & \texttt{ET} & \texttt{ET} & \texttt{ET} & 19.50 - 2 & \texttt{ET} & \texttt{ET} & \texttt{ET} \\
    \botrule
    \end{tabular}
    \end{table}

    \begin{table}[!h]
    \footnotesize
    \centering
    \caption{Cross-Domain Recall@8 Scores and Ranks}
    \label{tab:cross-domain-recall-scores-ranks-8}
    \begin{tabular}{l@{\hskip 0.15in}cccccccc}
    \toprule
    \textbf{Method} & \makecell{\textbf{WOS} \\-141} & \makecell{\textbf{NYT} \\-166} & \makecell{\textbf{EurLex} \\-3985} & \makecell{\textbf{EurLexDC} \\-410} & \makecell{\textbf{SciHTC} \\-83} & \makecell{\textbf{SciHTC} \\-800} & \makecell{\textbf{MIMIC3} \\-3681} & \makecell{\textbf{USPTO2M} \\-632} \\
    \midrule
    MatchXML & 93.42 - 5 & 78.53 - 2 & 54.08 - 2 & 95.50 - 3 & 83.87 - 5 & 67.32 - 1 & 38.82 - 3 & 88.08 - 3 \\
    XR-Transformer & 94.24 - 4 & 79.02 - 1 & 54.18 - 1 & 96.09 - 2 & 82.29 - 6 & 64.73 - 5 & 39.01 - 2 & 85.43 - 4 \\
    HILL & 95.83 - 1 & 74.47 - 5 & 53.21 - 3 & 96.13 - 1 & 79.13 - 8 & 59.69 - 7 & 33.24 - 4 & \texttt{ET} \\
    HiAGM-TP & 95.27 - 2 & 67.05 - 7 & 48.92 - 5 & 94.46 - 5 & 82.03 - 7 & 64.26 - 6 & \texttt{ET} & \texttt{ET} \\
    PLM-ICD & 94.97 - 3 & 78.18 - 3 & 51.30 - 4 & 94.47 - 4 & 86.11 - 1 & 66.85 - 3 & 44.03 - 1 & 90.54 - 2 \\
    FlatBERT & 67.38 - 7 & 68.51 - 6 & 19.65 - 7 & 19.14 - 7 & 84.71 - 4 & 65.83 - 4 & 5.19 - 5 & 91.03 - 1 \\
    THMM & 92.09 - 6 & 75.55 - 4 & 24.37 - 6 & 66.87 - 6 & 85.68 - 3 & 67.07 - 2 & \texttt{ET} & \texttt{ET} \\
    HR-SciBERT-mt & \texttt{ET} & \texttt{ET} & \texttt{ET} & \texttt{ET} & 85.92 - 2 & \texttt{ET} & \texttt{ET} & \texttt{ET} \\
    \botrule
    \end{tabular}
    \end{table}

    \begin{table}[!h]
    \footnotesize
    \centering
    \caption{Cross-Domain Precision@10 Scores and Ranks}
    \label{tab:cross-domain-precision-scores-ranks-10}
    \begin{tabular}{l@{\hskip 0.15in}cccccccc}
    \toprule
    \textbf{Method} & \makecell{\textbf{WOS} \\-141} & \makecell{\textbf{NYT} \\-166} & \makecell{\textbf{EurLex} \\-3985} & \makecell{\textbf{EurLexDC} \\-410} & \makecell{\textbf{SciHTC} \\-83} & \makecell{\textbf{SciHTC} \\-800} & \makecell{\textbf{MIMIC3} \\-3681} & \makecell{\textbf{USPTO2M} \\-632} \\
    \midrule
    MatchXML & 18.86 - 5 & 54.61 - 2 & 78.01 - 2 & 12.10 - 3 & 15.75 - 5 & 10.98 - 3 & 58.77 - 3 & 16.21 - 3 \\
    XR-Transformer & 19.01 - 4 & 55.01 - 1 & 78.14 - 1 & 12.17 - 2 & 15.43 - 7 & 10.58 - 5 & 58.92 - 2 & 15.55 - 4 \\
    HILL & 19.31 - 1 & 52.40 - 4 & 77.13 - 3 & 12.25 - 1 & 14.89 - 8 & 9.84 - 7 & 50.25 - 4 & \texttt{ET} \\
    HiAGM-TP & 19.20 - 2 & 45.79 - 7 & 70.91 - 5 & 11.98 - 5 & 15.60 - 6 & 10.58 - 5 & \texttt{ET} & \texttt{ET} \\
    PLM-ICD & 19.17 - 3 & 54.55 - 3 & 73.61 - 4 & 12.01 - 4 & 16.21 - 1 & 10.99 - 2 & 67.11 - 1 & 16.94 - 2 \\
    FlatBERT & 14.54 - 7 & 48.22 - 6 & 28.70 - 7 & 2.20 - 7 & 15.95 - 4 & 10.71 - 4 & 7.19 - 5 & 17.08 - 1 \\
    THMM & 18.63 - 6 & 52.08 - 5 & 35.56 - 6 & 8.52 - 6 & 16.14 - 2 & 11.06 - 1 & \texttt{ET} & \texttt{ET} \\
    HR-SciBERT-mt & \texttt{ET} & \texttt{ET} & \texttt{ET} & \texttt{ET} & 16.14 - 2 & \texttt{ET} & \texttt{ET} & \texttt{ET} \\
    \botrule
    \end{tabular}
    \end{table}

    \begin{table}[!h]
    \footnotesize
    \centering
    \caption{Cross-Domain Recall@10 Scores and Ranks}
    \label{tab:cross-domain-recall-scores-ranks-10}
    \begin{tabular}{l@{\hskip 0.15in}cccccccc}
    \toprule
    \textbf{Method} & \makecell{\textbf{WOS} \\-141} & \makecell{\textbf{NYT} \\-166} & \makecell{\textbf{EurLex} \\-3985} & \makecell{\textbf{EurLexDC} \\-410} & \makecell{\textbf{SciHTC} \\-83} & \makecell{\textbf{SciHTC} \\-800} & \makecell{\textbf{MIMIC3} \\-3681} & \makecell{\textbf{USPTO2M} \\-632} \\
    \midrule
    MatchXML & 94.29 - 5 & 83.87 - 2 & 62.67 - 2 & 95.74 - 3 & 86.62 - 5 & 70.39 - 3 & 43.90 - 3 & 89.34 - 3 \\
    XR-Transformer & 95.05 - 4 & 84.38 - 1 & 62.79 - 1 & 96.34 - 2 & 84.91 - 7 & 67.64 - 6 & 43.99 - 2 & 86.68 - 4 \\
    HILL & 96.53 - 1 & 79.70 - 5 & 61.91 - 3 & 96.72 - 1 & 82.17 - 8 & 63.50 - 7 & 37.26 - 4 & \texttt{ET} \\
    HiAGM-TP & 96.01 - 2 & 71.85 - 7 & 56.67 - 5 & 95.10 - 4 & 85.78 - 6 & 68.27 - 5 & \texttt{ET} & \texttt{ET} \\
    PLM-ICD & 95.84 - 3 & 83.58 - 3 & 59.03 - 4 & 95.07 - 5 & 89.23 - 1 & 70.82 - 2 & 50.25 - 1 & 92.13 - 2 \\
    FlatBERT & 72.70 - 7 & 72.93 - 6 & 22.13 - 7 & 19.99 - 7 & 87.89 - 4 & 69.73 - 4 & 5.19 - 5 & 92.61 - 1 \\
    THMM & 93.13 - 6 & 80.65 - 4 & 27.56 - 6 & 69.44 - 6 & 88.81 - 3 & 70.99 - 1 & \texttt{ET} & \texttt{ET} \\
    HR-SciBERT-mt & \texttt{ET} & \texttt{ET} & \texttt{ET} & \texttt{ET} & 88.83 - 2 & \texttt{ET} & \texttt{ET} & \texttt{ET} \\
    \botrule
    \end{tabular}
    \end{table}

\subsection{Dataset- and Method-Specific Patterns}\label{sec:dataset-method-specific-patterns}
\emph{Hierarchy depth} strongly influences the performance of certain methods. Shallow hierarchies (like in WOS-141) are better handled by HILL, HiAGM-TP, and THMM than datasets with deeper hierarchies (EurLex-*, SciHTC-*). \emph{Document length} and label structure are key factors because datasets with long documents and flat label structures (MIMIC3-3681, USPTO2M-632) seem to require specialized text processing approaches, which PLM-ICD handles effectively. Datasets with \emph{label sparsity}, i.e. with low average samples per label (EurLex-*), benefit from robust label-aware mechanisms, where MatchXML, XR-Transformer, and PLM-ICD excel. \emph{Sample abundance} can compensate for other complexities. For example, datasets with high sample counts per label (like SciHTC-83) enable even simpler methods to perform well, as demonstrated by THMM's unexpected success.

On the method side, the \emph{baseline} FlatBERT lags behind other methods but shows competitive results with sufficient training data (USPTO2M-632). \emph{Transformer-based} methods (MatchXML, XR-Transformer) excel with large label spaces but show limitations in exploiting label redundancy. 
\emph{Per node classification} methods face scalability issues, as HR-SciBERT-mt usually exceeds time limits.

\subsection{XR-Transformer Tree Initialization Variations}\label{sec:xr-tree-init}
The results (\ref{tab:xrt-precision-1}, \ref{tab:xrt-recall-1}) demonstrate that tree initialization methods have a small but consistent impact on XR-Transformer's performance. Random initialization consistently performs slightly worse than both PIFA and Gold methods across most datasets, with performance gaps of 0.1-0.5 percentage points in both precision and recall. Gold initialization generally achieves the best results (e.g., 97.67\% precision on EurLex-3985), followed closely by PIFA, while Random initialization ranks last. This suggests that leveraging either Gold or PIFA tree structures is beneficial, though the performance differences are modest.

\begin{table}[!htb]
    \tiny
    \centering
    \caption{XR-Transformer tree initialization variations precision@1}
    \label{tab:xrt-precision-1}
    \begin{tabular}{cccccccccc}
    \toprule
    Init Tree & WOS-141 & NYT-166 & EurLex-3956 & EurLex-3985  & EurLexDC & SciHTC-83 & SciHTC-800 & MIMIC3-3681 (512) & USPTO2M-632\\ \midrule
    PIFA & 86.49 & 96.64 & 87.83 & 97.51 & 89.80 & 61.53 & 40.69 & 88.32 & 80.99 \\ 
    Gold & 86.39 & 96.70 & 88.06 & 97.67 & 89.86 & 61.58 & 40.76 & 88.03 & 81.19\\ 
    Random &86.40 & 96.71  & 86.77 & 97.48 & 89.79 & 61.37 & 40.68 & 87.69 & 80.82 \\ 
    \botrule
    \end{tabular}
\end{table}

\begin{table}[!htb]
    \tiny
    \centering
    \caption{XR-Transformer tree initialization variations recall@1}
    \label{tab:xrt-recall-1}
    \begin{tabular}{cccccccccc}
    \toprule
    Init Tree & WOS-141 & NYT-166 & EurLex-3956 & EurLex-3985  & EurLexDC & SciHTC-83 & SciHTC-800 & MIMIC3-3681 (512) & USPTO2M-632 \\ \midrule
    PIFA & 43.25 & 22.93 & 17.82 & 8.09 & 78.91  & 36.29  & 27.43  & 7.37 & 54.75 \\
    Gold & 43.20 & 22.94 & 17.88 & 8.10 & 78.96 & 36.32 & 27.49  & 7.35 & 54.90 \\ 
    Random & 43.20 & 22.95 & 17.62 & 8.08 & 78.91 & 36.21 & 27.42 & 7.31 & 54.66 \\
    \botrule
    \end{tabular}
\end{table}

\end{appendices}
\end{document}